\documentclass[10pt,twocolumn,letterpaper]{article}

\usepackage{cvpr}
\usepackage{times}
\usepackage{epsfig}
\usepackage{graphicx}
\usepackage{amsmath}
\usepackage{amssymb,enumitem}

\usepackage{multirow}
\usepackage{tabls}
\usepackage{wrapfig}
\usepackage{hyperref}
\usepackage{amsmath,amssymb} 
\usepackage{amsmath,amssymb}
 \usepackage[ruled,vlined,linesnumbered,resetcount]{algorithm2e}
  \usepackage{booktabs}
 \usepackage[linesnumbered]{algorithm2e}
\usepackage{algorithmic}
\usepackage{caption}
\usepackage{subfigure}
\renewcommand{\algorithmicrequire}{ \textbf{Input:}} 
\renewcommand{\algorithmicensure}{ \textbf{Output:}} 
\usepackage[font=small,labelfont=bf]{caption}
   \usepackage{graphicx}
 \usepackage{booktabs}

\usepackage{epstopdf}
\usepackage{graphicx}
\usepackage{subfigure}
\usepackage{enumitem}
\usepackage{multirow}
\usepackage{tabls}
\usepackage{wrapfig}
\usepackage{hyperref}
\usepackage{amsmath,amssymb} 
\usepackage{amsmath,amssymb}
 \usepackage[ruled,vlined,linesnumbered,resetcount]{algorithm2e}
  \usepackage{booktabs}
 \usepackage[linesnumbered]{algorithm2e}
\usepackage{algorithmic}
\usepackage{caption}
\usepackage{subfigure}
\renewcommand{\algorithmicrequire}{ \textbf{Input:}} 
\renewcommand{\algorithmicensure}{ \textbf{Output:}} 
\usepackage[font=small,labelfont=bf]{caption}
   \usepackage{graphicx}
 \usepackage{booktabs}
\usepackage{epstopdf}
\usepackage{graphicx}
\usepackage{subfigure}
\usepackage{enumitem}

\cvprfinalcopy 

\begin{document}

\title{Density-aware Single Image De-raining using a Multi-stream  Dense Network}

\author{He  Zhang \qquad \qquad \qquad Vishal M. Patel \\
Department of Electrical and Computer Engineering\\Rutgers University, Piscataway, NJ 08854
\\
{\tt\small {\{he.zhang92,vishal.m.patel\}}@rutgers.edu}
}

    \setlength\abovedisplayskip{0pt}
    \setlength\belowdisplayskip{0pt}

\maketitle

\begin{abstract}
Single image rain streak removal is an extremely challenging problem due to the presence of non-uniform rain densities in images.   We present a novel density-aware multi-stream densely connected convolutional neural network-based algorithm, called DID-MDN, for joint rain density estimation and de-raining.    The proposed method enables the network itself to automatically determine the rain-density information and then efficiently remove the corresponding rain-streaks guided by the estimated rain-density label. To  better characterize rain-streaks with different scales and shapes,  a  multi-stream densely connected de-raining network is proposed which efficiently leverages features from different scales. Furthermore, a new dataset containing  images with rain-density labels is created and used to train the proposed density-aware network. Extensive experiments on synthetic and real datasets demonstrate that the proposed method achieves significant improvements over the  recent state-of-the-art methods.  In addition, an ablation study is performed to demonstrate the improvements obtained by different modules in the proposed method. Code can be found at: https://github.com/hezhangsprinter
\end{abstract}

\section{Introduction}
\label{sec:introduction}
In many applications such as drone-based video surveillance and self driving cars, one has to process images and videos containing undesirable artifacts such  as  rain,  snow,  and  fog.   Furthermore, the performance of many computer vision systems often degrades when they are presented with images containing some of these artifacts.  Hence, it is important to develop algorithms that can automatically remove these artifacts.   In this paper, we address the problem of rain streak removal from a single image.  Various methods have been proposed in the literature to address this problem \cite{rain_2016_gmm,derain_cvpr2017,derain_csc_17,dis_rain_2015,derain_lowrank,derain_tip12,derain_tip14,derain_depth_sparse,derain_2017_zhang,derain_cvpr2017_multi,derain_tip17}.

\begin{figure}[t]
	\centering
	\begin{minipage}{.155\textwidth}
		\centering
		\includegraphics[width=2.9cm,height=2.5cm]{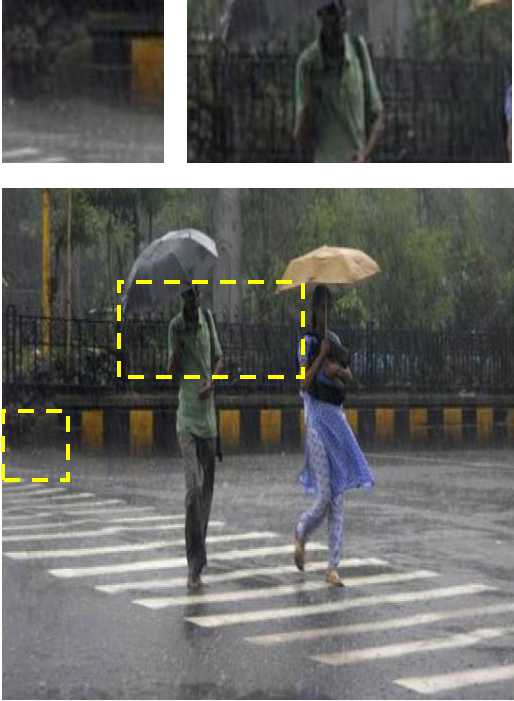}
		\captionsetup{labelformat=empty}
		\captionsetup{justification=centering}
		\caption*{(a)}
	\end{minipage}
	\begin{minipage}{.155\textwidth}
		\centering
		\includegraphics[width=2.9cm,height=2.5cm]{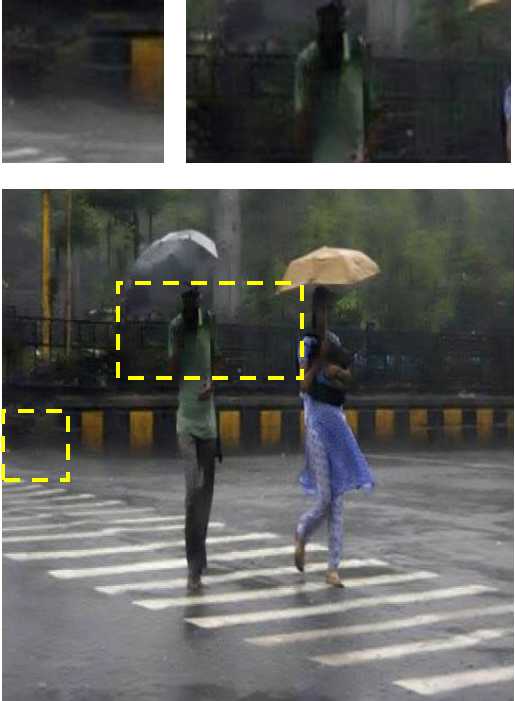}
		\captionsetup{labelformat=empty}
		\captionsetup{justification=centering}
		\caption*{(b)}
	\end{minipage}
	\begin{minipage}{.155\textwidth}
		\centering
		\includegraphics[width=2.9cm,height=2.5cm]{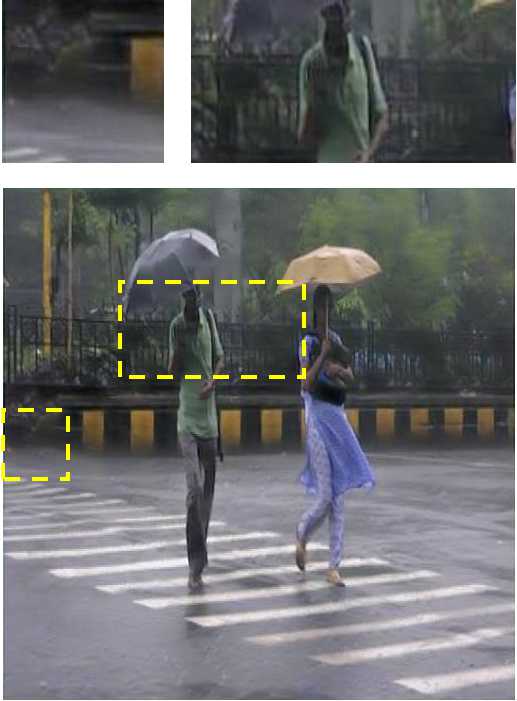}
		\captionsetup{labelformat=empty}
		\captionsetup{justification=centering}
		\caption*{(c)}
	\end{minipage}	\\
	\begin{minipage}{.155\textwidth}
		\centering
		\includegraphics[width=2.75cm,height=1.80cm]{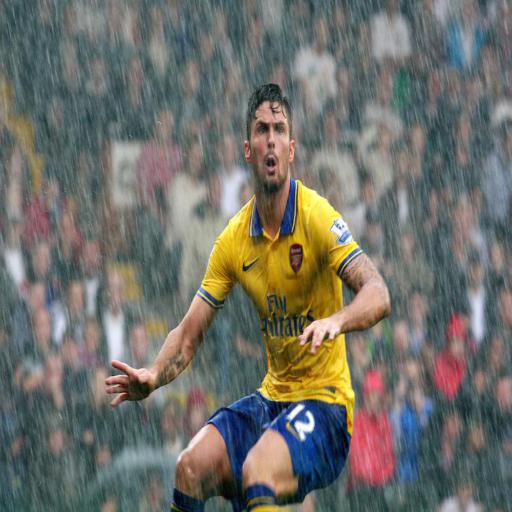}
		\captionsetup{labelformat=empty}
		\captionsetup{justification=centering}
		\caption*{(d)}
	\end{minipage}
	\begin{minipage}{.155\textwidth}
		\centering
		\includegraphics[width=2.75cm,height=1.80cm]{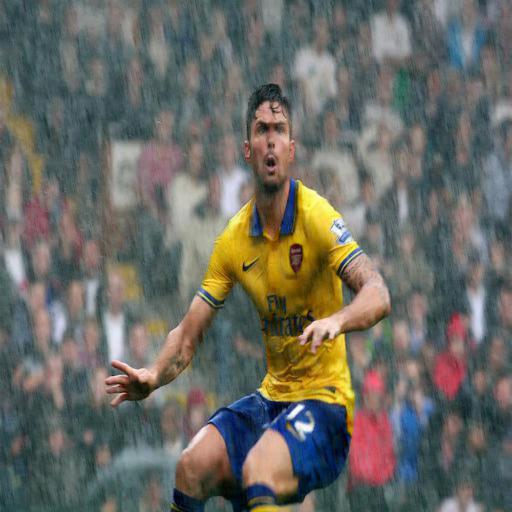}
		\captionsetup{labelformat=empty}
		\captionsetup{justification=centering}
		\caption*{(e)}
	\end{minipage}
	\begin{minipage}{.155\textwidth}
		\centering
		\includegraphics[width=2.75cm,height=1.80cm]{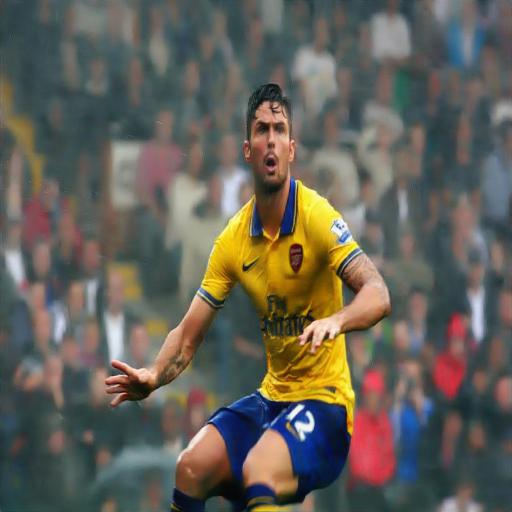}
		\captionsetup{labelformat=empty}
		\captionsetup{justification=centering}
		\caption*{(f)}
	\end{minipage}
	\vskip -8pt\caption{Image de-raining results.  (a) Input rainy image. (b) Result from Fu \emph{et al}. \cite{derain_cvpr2017}. (c) DID-MDN.   (d) Input rainy image. (e) Result from Li \emph{et al}. \cite{derain_cvpr2017_multi}. (f) DID-MDN. Note that \cite{derain_cvpr2017} tends to over de-rain the image while \cite{derain_cvpr2017_multi} tends to under de-rain the image.  
	 \label{fig:motivation}}
\end{figure}

One of the main limitations of the existing single image de-raining methods is that they are designed to deal with certain types of rainy images and they do not effectively consider various shapes, scales and density of rain drops into their algorithms.   State-of-the-art de-raining algorithms such as \cite{derain_cvpr2017_multi,derain_cvpr2017} often tend to over de-rain or under de-rain the image if the rain condition present in the test image is not properly considered during training. 
For example, when a rainy image shown in Fig.~\ref{fig:motivation}(a) is de-rained using the method of Fu \emph{et al.} \cite{derain_cvpr2017}, it tends to remove some important parts  in the de-rained image such as the right arm of the person, as shown in Fig.~\ref{fig:motivation}(b).  Similarly, when \cite{derain_cvpr2017_multi} is used to de-rain the image shown in Fig.~\ref{fig:motivation}(d), it tends to under de-rain the image and leaves some rain streaks in the output de-rained image. Hence, more adaptive and efficient methods, that can deal with different rain density levels present in the image, are needed.    

One possible solution to this problem is to build a very large training dataset with sufficient rain conditions containing various rain-density levels with different orientations and scales.    This has been achieved by Fu \emph{et al}. \cite{derain_cvpr2017} and Yang \emph{et al.}\cite{derain_cvpr2017_multi}, where they synthesize a  novel  large-scale dataset consisting of rainy images with various conditions and they train a single network based on this dataset for image de-raining.  
 However, one drawback of this approach is that a single network may not be capable enough  to learn all types of variations present in the training samples.  It can be observed from Fig.~\ref{fig:motivation} that both methods tend to either over de-rain or under de-rain results.  Alternative solution to this problem is to  learn a density-specific model for de-raining. However,  this solution lacks flexibility in practical de-raining as the density label information is needed  for a given rainy image to determine which network to choose for de-raining.

\begin{figure*}[t]
	\centering
	\begin{minipage}{0.90\textwidth}
		\includegraphics[width=1\textwidth]{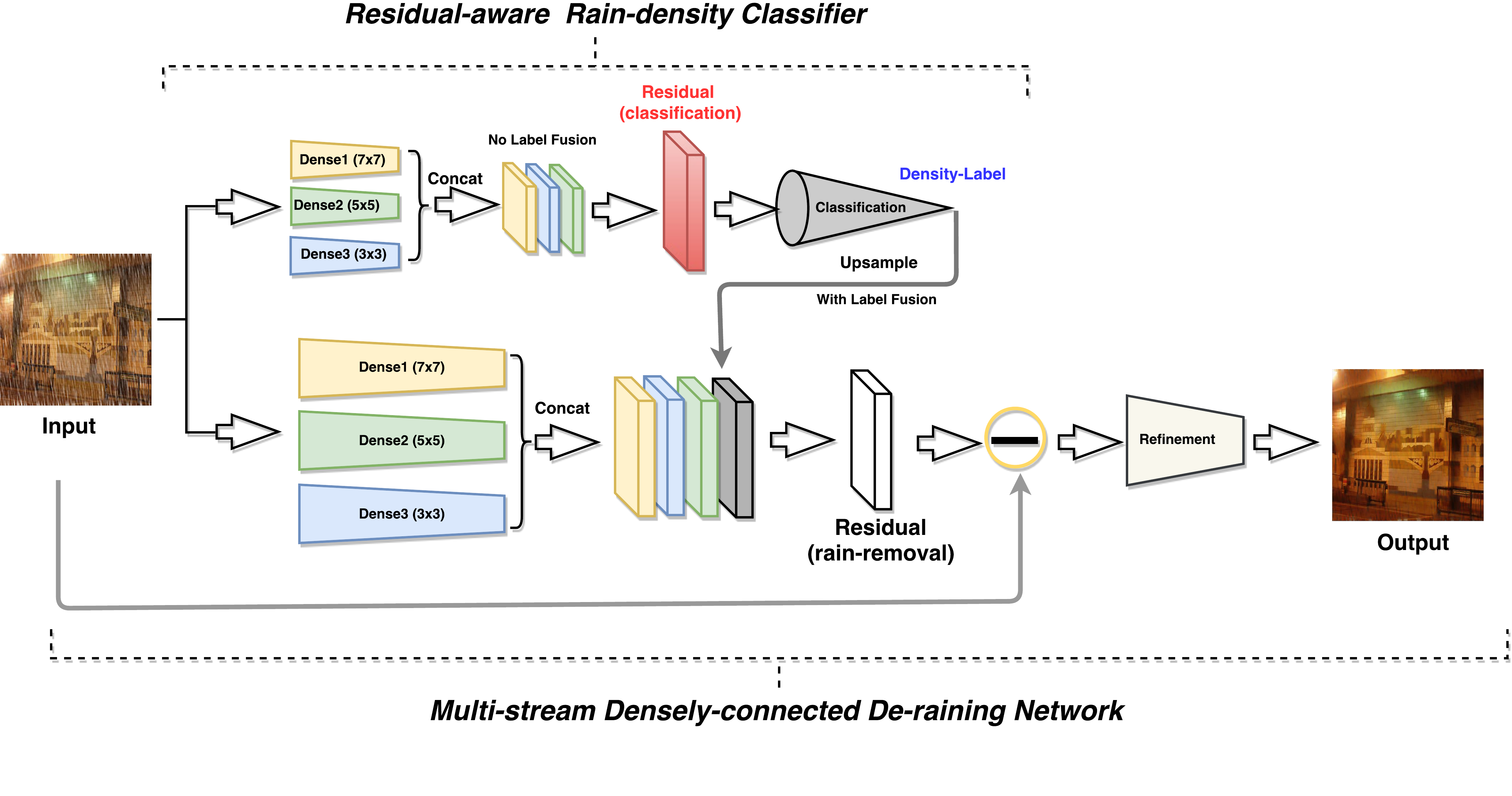}
		\captionsetup{labelformat=empty}
		\captionsetup{justification=centering}
	\end{minipage}
	\vskip-20pt
	\caption{An overview of the proposed DID-MDN method. The proposed network contains two modules: (a) residual-aware rain-density classifier, and (b) multi-stream densely-connected de-raining network.  The goal of the residual-aware rain-density classifier is to determine the rain-density level given a rainy image.  On the other hand, the multi-stream densely-connected de-raining network is designed to efficiently remove the rain streaks from the rainy images guided by the  estimated rain-density information.} 
	\label{fig:overview}
\end{figure*}

In order to address these issues, we propose a novel Density-aware Image De-raining method using a Multi-stream Dense Network (DID-MDN) that can automatically determine the rain-density information (i.e. heavy, medium or light) present in the input image (see Fig.~\ref{fig:overview}).   The proposed method consists of two main stages: rain-density classification and rain streak removal.  To accurately estimate the  rain-density  level,  a new  residual-aware classifier that makes use of the residual component in the rainy image for density classification is proposed in this paper.  The rain streak removal algorithm is based on a multi-stream densely-connected network that takes into account the distinct scale and shape information of rain streaks.  Once the rain-density level is estimated, we fuse the estimated density information into our final multi-stream densely-connected network to get the final de-rained output.  Furthermore, to efficiently train the proposed network, a large-scale dataset consisting of 12,000 images with different  rain-density levels/labels (i.e. heavy, medium and light) is synthesized.  Fig.~\ref{fig:motivation}(c) \& (d) present sample results from our network, where one can clearly see that DID-MDN does not over de-rain or under de-rain the image and is able to provide better results as compared to \cite{derain_cvpr2017} and \cite{derain_cvpr2017_multi}.

This paper makes the following contributions:
\begin{itemize}[noitemsep]     
\item[1.]  A novel DID-MDN method which automatically determines the rain-density information and then efficiently removes the corresponding rain-streaks guided by the estimated rain-density label is proposed. 

\item[2.] Based on the observation that residual can be used as a better feature representation in characterizing the rain-density information,  a novel residual-aware classifier to efficiently determine the density-level of a given rainy image is proposed in this paper.
\item[3.] A new synthetic dataset consisting of 12,000 training images with rain-density labels and 1,200 test images is synthesized. To the best of our knowledge, this is the first dataset that contains the rain-density label information.  Although the network is trained on our synthetic dataset,  it generalizes well to real-world rainy images.
\item[4.] Extensive experiments are conducted on three highly
challenging  datasets  (two synthetic and one real-world)  and  comparisons
are  performed  against  several  recent  state-of-the-art
approaches. Furthermore, an ablation study is conducted to
demonstrate the effects of different modules in the proposed network.
\end{itemize}

\section{Background and Related Work}\label{sec:related}
In this section, we briefly review several recent related
works on single image de-raining and multi-scale feature aggregation.  

\subsection{Single Image De-raining}
Mathematically, a rainy image $\mathbf{y}$ can be modeled as  a linear combination of a rain-streak component $\mathbf{r}$ with a clean background image $\mathbf{x}$, as follows
\begin{equation}\label{eq:observation}
\mathbf{y}=\mathbf{x}+\mathbf{r}.
\end{equation}
In single image de-raining, given $\mathbf{y}$ the goal is to recover $\mathbf{x}$.   As can be observed from \eqref{eq:observation} that image de-raining is a highly ill-posed problem. Unlike video-based methods \cite{derain_cvpr17_video,derain_iccv17_video,derain_video_ijcv}, which leverage temporal information in removing rain components,  prior-based methods have been proposed in the literature to deal with this problem.  These include sparse coding-based methods \cite{derain_tip12,derain_tip14, derain_iccv17}, low-rank representation-based methods \cite{derain_lowrank,derain_csc_17} and GMM-based (gaussian mixture model) methods \cite{rain_2016_gmm}.  One of the limitations of some of these prior-based methods is that they often tend to over-smooth the image details \cite{derain_tip12,derain_csc_17}.

Recently, due to the immense success of deep learning in both high-level and low-level vision tasks \cite{deep_residue,tao_stackgan_cvpr2018, zizhao_mdnet,peng_iccv17, jia_differ, style_hang},  several CNN-based methods have also been proposed for image de-raining \cite{cnn_derain2,derain_tip17,derain_cvpr2017_multi,derain_cvpr2017}.   In these methods, the idea is to learn a mapping  between input rainy images and their corresponding ground
truths using a CNN structure.  

\subsection{Multi-scale Feature Aggregation}
It has been observed that combining convolutional features at different levels (scales) can lead to a better representation of an object in the image and its surrounding context \cite{spp,psp_net,deep_residue,dense_net}.    For instance, to efficiently leverage features obtained from different scales, the FCN (fully convolutional network) method \cite{fcn} uses skip-connections and adds high-level prediction layers to intermediate layers to generate pixel-wise prediction results at multiple resolutions.  Similarly, the U-Net architecture \cite{unet} consists of a contracting path to capture the context and a symmetric expanding path that enables the precise localization.  The HED model \cite{deep_supervision} employs deeply supervised structures, and automatically learns rich hierarchical representations that are fused to resolve the challenging ambiguity in edge and object boundary detection.   Multi-scale features have also been leveraged in various applications such as semantic segmentation \cite{psp_net}, face-alignment \cite{peng_face}, visual tracking \cite{kunpeng_ijcai} crowd-counting \cite{crowd_counting_Vishwanath}, action recognition \cite{yizhu_hiddentwo}, depth estimation \cite{depth_nips_14}, single image dehazing \cite{dehaze_2016_eccv,dehaze_2017_joint} and also in single image de-raining \cite{derain_cvpr2017_multi}.  Similar to \cite{derain_cvpr2017_multi}, we also leverage a multi-stream network to capture the rain-streak components with different scales and shapes.  However, rather than using two convolutional layers with different dilation factors to combine features from different scales, we leverage the densely-connected block \cite{dense_net} as the building module and then we connect features from each block together for the final rain-streak removal. The ablation study demonstrates the effectiveness of our proposed network compared with the structure proposed in \cite{derain_cvpr2017_multi}. 

\section{Proposed Method}
The proposed DID-MDN architecture mainly consists of two modules: (a) residual-aware rain-density classifier, and (b) multi-stream densely connected de-raining network.  The residual-aware rain-density classifier aims to determine the rain-density level given a rainy image. On the other hand, the multi-stream densely connected de-raining network is designed to efficiently remove the rain streaks from the rainy images guided by the  estimated rain-density information.    The entire network architecture of the proposed DID-MDN method is shown in Fig.~\ref{fig:overview}.

\subsection{Residual-aware Rain-density Classifier}
As discussed above, even though some of the previous methods achieve significant improvements on the de-raining performance, they often tend to over de-rain or under de-rain the image.   This is mainly due to the fact that a single network may not be sufficient enough to learn different rain-densities occurring in practice.  We believe that incorporating density level information into the network can benefit the overall learning procedure and hence can guarantee better generalization to different rain conditions \cite{derain_cvpr17_video}.  Similar observations have also been made in \cite{derain_cvpr17_video}, where they use two different  priors to characterize light rain  and heavy rain, respectively.  Unlike using two priors to characterize different rain-density conditions \cite{derain_cvpr17_video}, the rain-density label estimated from a CNN classifier is used for guiding the de-raining process.  To  accurately estimate the density information given a rainy input image, a residual-aware rain-density classifier is proposed, where the residual information is leveraged  to better represent the rain features. In addition, to train the classier, a large-scale synthetic dataset  consisting of 12,000 rainy images with density labels is synthesized.  Note that there are only three types of classes (i.e. labels) present in the dataset and they correspond to low, medium and high density.

One common strategy in training a new classifier is to fine-tune a pre-defined  model such as VGG-16 \cite{vgg}, Res-net \cite{deep_residue} or Dense-net \cite{dense_net} on the newly introduced dataset. One of  the fundamental reasons to leverage a fine-tune strategy for the new dataset is that discriminative features encoded in these pre-defined models can be beneficial in accelerating the training  and it can also guarantee better generalization.  However, we observed that directly fine-tuning such a `deep' model on our task is not an efficient solution.   This is mainly due to the fact that high-level features (deeper part) of a CNN tend to pay more attention to localize the discriminative objects in the input image \cite{feature_vis_bolei}.  Hence, relatively small rain-streaks may not be localized well in these high-level features. In other words, the rain-streak information may be lost in the high-level features and hence may degrade the overall classification performance. As a result, it is important to come up with a better feature representation to effectively characterize rain-streaks (i.e. rain-density).

From \eqref{eq:observation}, one can regard $\mathbf{r}=\mathbf{y}-\mathbf{x}$ as the residual component which can be used to characterize the rain-density.   To estimate the residual component ($\mathbf{\hat{r}}$) from the observation $\mathbf{y}$,  a multi-stream dense-net (without the label fusion part) using the new dataset with heavy-density is trained.  Then, the estimated residual is regarded as the input to train the final classifier.  In this way, the residual estimation part can be regarded as the feature extraction procedure~\footnote{Classificaiton network can be regarded as two parts: 1.Feature extractor and 2. Classifer}, which is discussed in Section 3.2.  The classification part is mainly composed of three convolutional layers (Conv) with kernel size $3\times3$, one average pooling (AP) layer with kernel size 9$\times$9 and  two fully-connected layers (FC).  Details of the classifier are as follows:\\
\emph{Conv(3,24)-Conv(24,64)-Conv(64,24)-AP- FC(127896,512)-FC(512,3)},\\
where (3,24) means that the input consists of 3 channels and the output consists of 24 channels.  Note that the final layer  consists  of  a  set  of  3  neurons indicating  the  rain-density  class
of  the  input  image (i.e. low, medium, high). An ablation study, discussed in Section 4.3, is conducted to demonstrate the effectiveness of proposed residual-aware classifier as compared with the VGG-16 \cite{vgg} model.
\\


\noindent {\bf{Loss for the Residual-aware Classifier:}}. To efficiently train the classifier, a two-stage training protocol is leveraged. A residual  feature extraction network is firstly trained to estimate the residual part of the given rainy image, then a classification sub-network is trained using the estimated residual as the input and is optimized via the ground truth labels (rain-density). Finally, the two stages (feature extraction and classification) are jointly optimized. The overall loss function used to train the residual-aware classier is as follows:  
 \begin{equation}
 \label{eq:loss_cla}
 \begin{split}
 \mathcal{L}= \mathcal{L}_{E,r}+  \mathcal{L}_C,
 \end{split}
  \end{equation}
where $\mathcal{L}_{E,r}$ indicates the per-pixel Euclidean-loss to estimate the residual component and $\mathcal{L}_C$ indicates the cross-entropy loss for rain-density classification.
 
\begin{figure}[t]
	\centering
	\begin{minipage}{.23\textwidth}
		\centering
		\includegraphics[width=4.1cm,height=3.5cm]{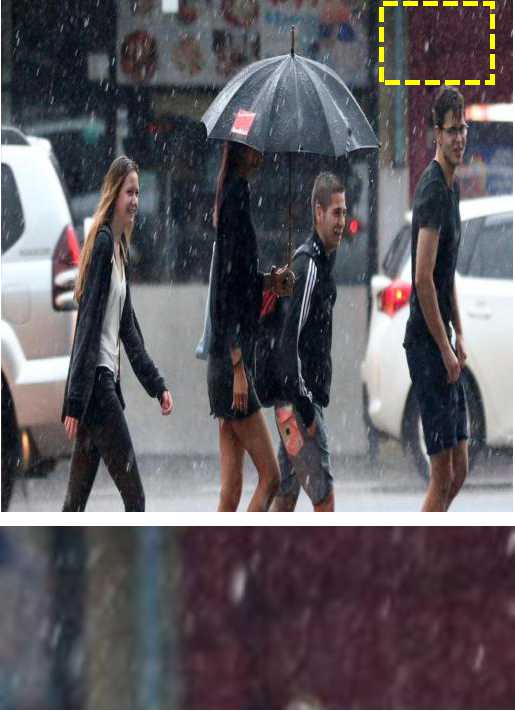}
		\captionsetup{labelformat=empty}
		\captionsetup{justification=centering}
		\caption*{(a)}
	\end{minipage}
	\begin{minipage}{.23\textwidth}
		\centering
		\includegraphics[width=4.1cm,height=3.5cm]{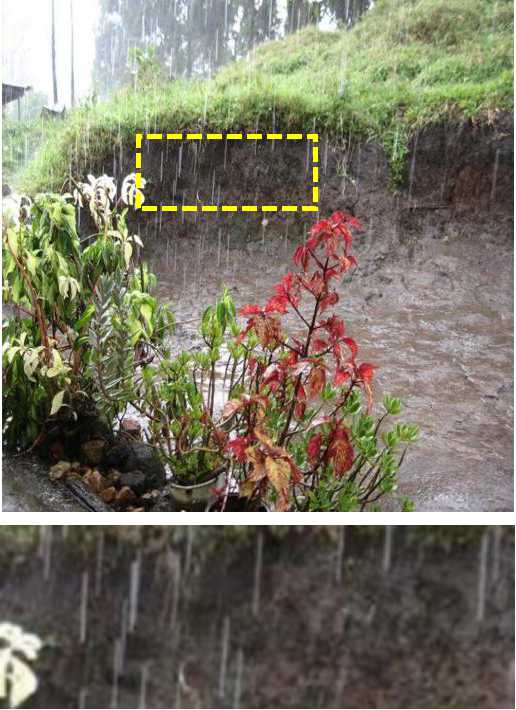}
		\captionsetup{labelformat=empty}
		\captionsetup{justification=centering}
    	\caption*{(b)}
	\end{minipage}	
		\vskip-10pt \caption{Sample images containing rain-streaks with various scales and shapes.(a) contains smaller rain-streaks, (b) contains longer rain-streaks.} \label{fig:moti_scale}
\end{figure}

\subsection{Multi-stream Dense Network}
It is well-known that different rainy images contain rain-streaks with different scales and shapes.   Considering the images shown in Fig.~\ref{fig:moti_scale},    the rainy image in Fig.~\ref{fig:moti_scale} (a) contains smaller rain-streaks, which can be captured by small-scale features (with smaller receptive fields), while the image in Fig.~\ref{fig:moti_scale} (b) contains longer rain-streaks, which can be captured by large-scale features (with larger receptive fields).  Hence, we believe that  combining features from different scales can be a more efficient way to capture various rain streak components \cite{derain_context,derain_cvpr2017_multi}.

Based on this observation and motivated by the success of using multi-scale features for single image de-raining \cite{derain_cvpr2017_multi},  a more efficient multi-stream  densely-connected network to estimate the rain-streak components is proposed, where each stream is built on the dense-block introduced in \cite{dense_net} with different kernel sizes (different receptive fields).   These multi-stream blocks are denoted by Dense1 $(7\times7)$, Dense2 $(5\times5)$, and Dense3 $(3\times3)$, in yellow, green and blue blocks, respectively in Fig.~\ref{fig:overview}.   In addition, to further improve the information flow among different blocks and to leverage features from each dense-block in estimating the rain streak components,  a modified connectivity is introduced, where all the features from each block are concatenated together for rain-streak estimation. Rather than leveraging only two convolutional layers in each stream \cite{derain_cvpr2017_multi}, we create short paths among features from different scales to strengthen feature aggregation and to obtain better convergence. To demonstrate the effectiveness of  our proposed  multi-stream network compared with the multi-scale structure proposed in \cite{derain_cvpr2017_multi},  an ablation study is conducted, which is described in Section~\ref{sec:exp}. 

To leverage the rain-density information to guide the de-raining process,  the up-sampled label map \footnote{For example, if the label is 1, then the corresponding up-sampled label-map is of the same dimension as the output features from each stream and all the pixel values of  the label map are 1.} is concatenated with the rain streak features from all three streams. Then, the concatenated features are used to estimate the residual ($\mathbf{\hat{r}}$) rain-streak information. In addition, the residual is subtracted from the input rainy image to estimate the coarse de-rained image. Finally, to further refine the estimated coarse de-rained image and make sure better details well preserved, another two convolutional layers with ReLU are adopted as the final refinement.

\begin{figure}[htp!]
 	\centering
 	\begin{minipage}{0.45\textwidth}
 		\includegraphics[width=1\textwidth]{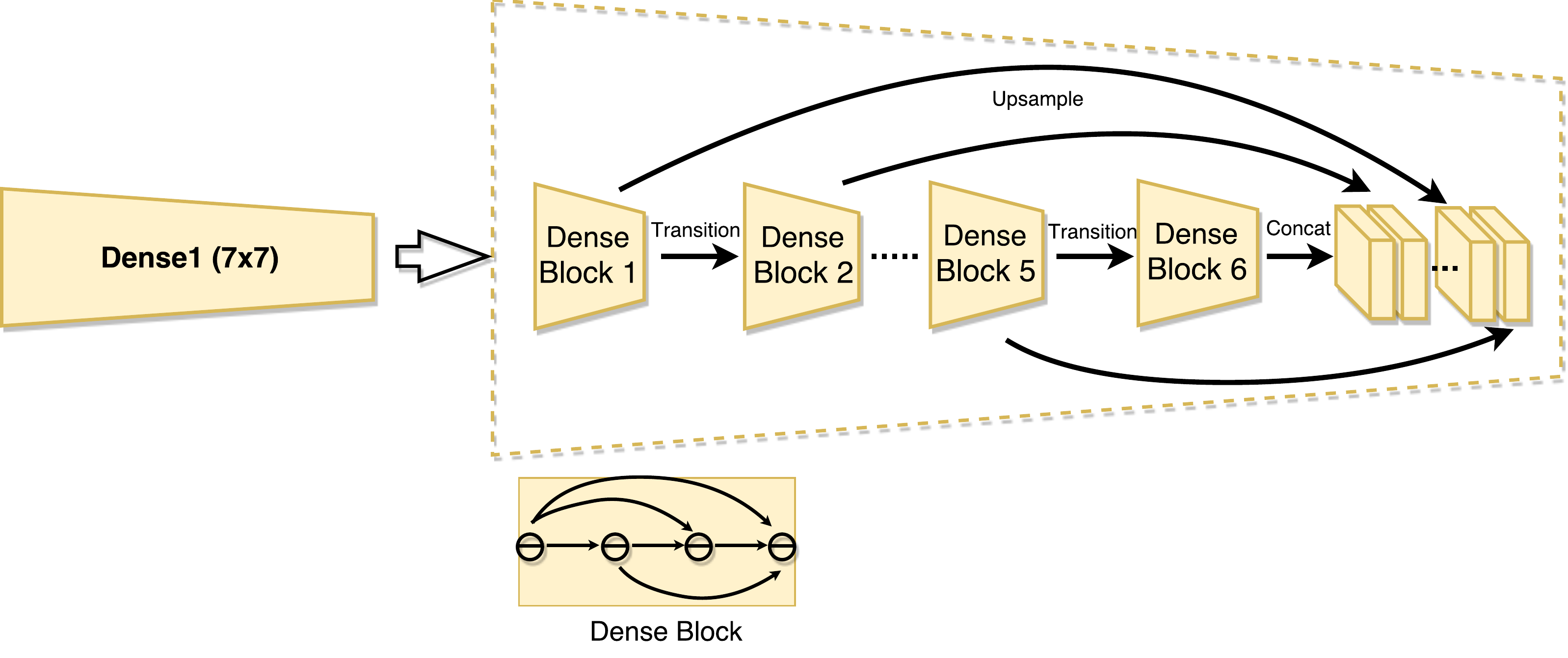}
 		\captionsetup{labelformat=empty}
 		\captionsetup{justification=centering}
 	\end{minipage}
 	\vskip-10pt
 	\caption{Details of the first stream \emph{Dense1}.} 
 	\label{fig:multi_column}
 \end{figure}

There are six dense-blocks in  each stream. Mathematically, each stream can be represented as
\begin{equation}\label{eq:single_stream}
\mathbf{s}_j={cat}[DB_1, DB_2,..., DB_6],
\end{equation}
where $cat$ indicates concatenation, $DB_i, i=1, \cdots 6$ denotes the output from the $i$th dense block, and $\mathbf{s}_j, j=1, 2, 3$ denotes the $j$th stream. Furthermore, we adopt different transition layer combinations\footnote{The transition layer can function as up-sample transition, down-sample transition or no-sampling transition \cite{dense_fully}.} and kernel sizes in each stream.  Details of each stream are as follows:\\
 \noindent \textbf{Dense1}: three transition-down layers, three transition-up layers and kernel size $7\times7$.\\ 
 \noindent \textbf{Dense2}: two transition-down layers, two no-sampling transition layers, two transition-up layers and kernel size $5\times5$.\\
 \noindent \textbf{Dense3}: one transition-down layer, four no-sampling transition layers, one transition-up layer and kernel size $3\times3$.\\
 Note that each dense-block is followed by a transition layer.   Fig~\ref{fig:multi_column} presents an overview of the first stream, \textbf{Dense1}.\\
 
\noindent {\bf{Loss for the De-raining Network:}}. Motivated by the observation that CNN feature-based loss can better improve the semantic edge information  \cite{perceptual_loss,SR_photorea} and to further enhance the visual quality of the estimated de-rained image \cite{derain_2017_zhang}, we also leverage a weighted combination of pixel-wise Euclidean loss and the feature-based loss. The loss for training the multi-stream densely connected network is as follows
 \begin{equation}
 \label{eq:loss_all}
 \begin{split}
 \mathcal{L}= \mathcal{L}_{E,r}+ \mathcal{L}_{E,d}+  \lambda_F\mathcal{L}_F,
 \end{split}
  \end{equation}
where $\mathcal{L}_{E,d}$ represents the per-pixel Euclidean loss function to reconstruct the de-rained image and  $\mathcal{L}_F$ is the feature-based loss for the de-rained image, defined as
 \begin{equation}\label{eq:perc_loss}
 \begin{split}
 \mathcal{L}_F= \frac{1}{CWH} \|F(\mathbf{\hat{x}})^{c,w,h}-F
 (\mathbf{x})^{c,w,h}\|_2^2,
 \end{split}
 \end{equation}
 where $F$ represents a non-linear CNN transformation and $\hat{\mathbf{x}}$ is the recovered de-rained image. Here, we have assumed that the features are of size $w\times h$ with $c$  channels.  In our method, we compute the feature loss from the layer relu1$\_$2 of the VGG-16 model \cite{vgg}.
 
\begin{table*}[htp!]
	\centering
	\caption{Quantitative results evaluated in terms of average SSIM and PSNR (dB) (SSIM/PSNR).}
	\resizebox{0.95\textwidth}{!}{%
\begin{tabular}{|c|c|c|c|c|c|c|c|c|}
	\hline
	& {Input} & {DSC \cite{dis_rain_2015} (ICCV'15)} & {GMM \cite{rain_2016_gmm}} (CVPR'16) & {CNN \cite{derain_tip17}} (TIP'17) & JORDER \cite{derain_cvpr2017_multi} (CVPR'17) & {DDN \cite{derain_cvpr2017}} (CVPR'17) & JBO \cite{derain_iccv17} (ICCV'17) & DID-MDN \\ \hline\hline
	\emph{Test1} & 0.7781/21.15 & 0.7896/21.44 & 0.8352/22.75 & 0.8422/22.07  & 0.8622/24.32 & 0.8978/ 27.33 & 0.8522/23.05 & \textbf{0.9087}/ \textbf{27.95} \\ \hline
	\emph{Test2} & 0.7695/19.31 & 0.7825/20.08 & 0.8105/20.66 & 0.8289/19.73 & 0.8405/22.26 & 0.8851/25.63 & 0.8356/22.45 & \textbf{0.9092}/   \textbf{26.0745}	 \\ \hline
\end{tabular}
	}	\label{ta:quantitive}
\end{table*}
 


\subsection{Testing}
During testing, the rain-density label information using the proposed residual-aware classifier  is estimated. Then,  the up-sampled label-map with the corresponding input image are fed into the multi-stream network to get the final de-rained image.  


\section{Experimental Results}
\label{sec:exp}
In this section, we present the  experimental  details and  evaluation  results  on both synthetic and real-world datasets.   De-raining performance on the synthetic data is evaluated in terms of PSNR and SSIM \cite{ssim}.  Performance of different methods on real-world images is evaluated visually since the ground truth images are not available.   The proposed DID-MDN method is compared  with the following recent state-of-the-art methods: (a) Discriminative sparse coding-based method (DSC) \cite{dis_rain_2015} (ICCV'15), (b) Gaussian mixture model (GMM) based method \cite{rain_2016_gmm} (CVPR'16), (c) CNN method (CNN) \cite{derain_tip17}  (TIP'17), (d) Joint Rain Detection and Removal (JORDER) method \cite{derain_cvpr2017_multi} (CVPR'17), (e) Deep detailed Network method (DDN) \cite{derain_cvpr2017} (CVPR'17), and (f) Joint Bi-layer Optimization (JBO) method \cite{derain_iccv17} (ICCV'17).

\subsection{Synthetic Dataset}
\begin{figure}[htp!]
 	\centering
 	\begin{minipage}{.155\textwidth}
 		\centering
 		\includegraphics[width=2.7cm,height=1.5cm]{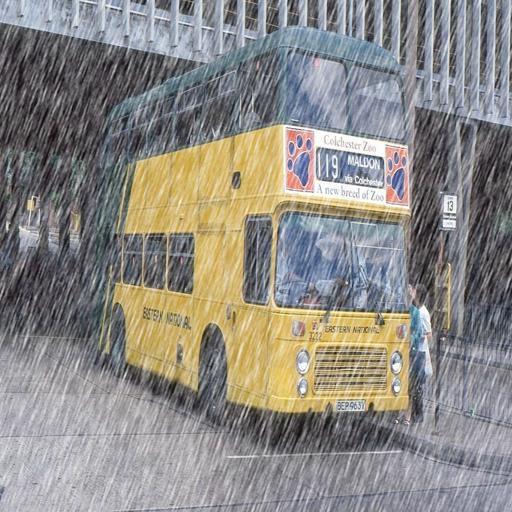}
 		\captionsetup{labelformat=empty}
 		\captionsetup{justification=centering}
 		\vskip-10pt
 		\caption*{Heavy} 
 	\end{minipage}
 	\begin{minipage}{.155\textwidth}
 		\centering
 		\includegraphics[width=2.7cm,height=1.5cm]{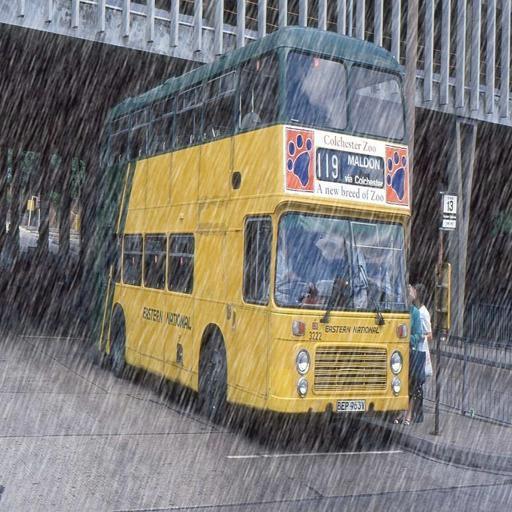}
 		\captionsetup{labelformat=empty}
 		\captionsetup{justification=centering}
 		\vskip-10pt
 		\caption*{Medium} 
 	\end{minipage}
 	\begin{minipage}{.155\textwidth}
 		\centering
 		\includegraphics[width=2.7cm,height=1.5cm]{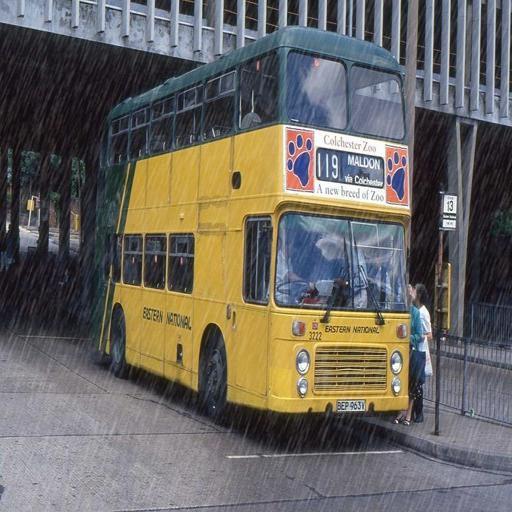}
 		\captionsetup{labelformat=empty}
 		\captionsetup{justification=centering}
 		\vskip-10pt
 		\caption*{Light} 
 	\end{minipage}
 	\vskip -10pt\caption{Samples synthetic images in three different conditions.} \label{fig:sample}
 \end{figure}

Even though there exist several large-scale synthetic datasets  \cite{derain_cvpr2017,derain_2017_zhang,derain_cvpr2017_multi}, they lack the availability of  the  corresponding rain-density label information for each synthetic rainy image.   Hence, we develop a new dataset, denoted as \emph{Train1}, consisting of 12,000 images, where each image is assigned a label  based on its corresponding rain-density level.  There are three rain-density labels present in the dataset (e.g. light, medium and heavy).  There are roughly 4,000 images per rain-density level in the dataset. Similarly, we also synthesize a new test set, denoted as \emph{Test1}, which consists of a total of 1,200 images.   It is ensured that each dataset contains rain streaks with different orientations  and scales.   Images are synthesized using Photoshop.   We modify the noise level introduced in step 3  of \footnote{http://www.photoshopessentials.com/photo-effects/photoshopweather-effects-rain/}
to generate different rain-density images, where light, medium and heavy rain conditions correspond to the noise levels $5\%\sim35\%$, $35\%\sim65\%$, and $65\%\sim95\%$, respectively \footnote{The reason why we use three labels is that during our experiments, we found that having more than three rain-density levels does not significantly improve the performance. Hence, we only use  three labels (heavy, medium and light) in the experiments.}.   Sample synthesized images under these three conditions are shown in Fig~\ref{fig:sample}.   To better test the  generalization capability of the proposed method, we  also randomly sample 1,000 images from the  synthetic dataset provided by Fu~\cite{derain_cvpr2017} as another testing set, denoted as \emph{Test2}. 

\subsection{Training Details}
 During training, a 512 $\times$ 512 image is randomly cropped from the input image (or its horizontal flip) of size 586$\times$586. Adam  is used as optimization algorithm with a mini-batch size of 1. The learning rate starts from 0.001 and is divided by 10 after 20 epoch.  The models are trained for up to 80$\times$12000 iterations. We use a weight decay of 0.0001 and a momentum of 0.9. The entire network is trained  using the Pytorch framework. During training, we set $\lambda_F=1$. All the parameters are defined via cross-validation using the validation set.

\begin{table}
	\centering
	\caption{Quantitative results compared with three baseline configurations on \emph{Test1}.}
	\resizebox{0.40\textwidth}{!}{%
\begin{tabular}{|c|c|c|c|c|}
	\hline
	& Single & Yang-Multi \cite{derain_cvpr2017_multi}  & Multi-no-label & DID-MDN \\ \hline\hline
	PSNR (dB) & 26.05 & 26.75  & {27.56} & \textbf{27.95} \\ \hline
	SSIM & 0.8893 & 0.8901 & {0.9028} & \textbf{0.9087} \\ \hline
\end{tabular}}
	\label{tab:baselinetable}
\end{table}

\subsection{Ablation Study}
The first ablation study is conducted to demonstrate the effectiveness of the proposed residual-aware classifier compared to the VGG-16 \cite{vgg} model. The two classifiers are trained using  our synthesized training samples \emph{Train1} and tested on the \emph{Test1} set. The classification accuracy corresponding to both classifiers on \emph{Test1} is tabulated in Table~\ref{tab:accuracy}. It can be observed that the proposed residual-aware classifier is more accurate than the VGG-16 model for predicting the rain-density levels.
\begin{table}
	\centering
	\caption{Accuracy of rain-density estimation evaluated on \emph{Test1}.}
	\resizebox{0.25\textwidth}{!}{%
\begin{tabular}{|c|c|c|c|c|}
	\hline
	& VGG-16 \cite{vgg} & Residual-aware \\ \hline\hline
	Accuracy & 73.32 \% & 85.15 \%  \\ \hline
\end{tabular}}
	\label{tab:accuracy}
\end{table}

In the second ablation study, we demonstrate the effectiveness of different modules in our method by conducting the following experiments: 
\begin{itemize}[noitemsep]   
  \item \textbf{Single}: A single-stream densely connected network (\textbf{Dense2}) without the procedure of label fusion. 
  \item \textbf{Yang-Multi  \cite{derain_cvpr2017_multi}\footnote{To better demonstrate the effectiveness of our proposed muli-stream network compared with the state-of-the-art multi-scale structure proposed in \cite{derain_cvpr2017_multi}, we replace our multi-stream dense-net part with the multi-scale structured in \cite{derain_cvpr2017_multi} and keep all the other parts the same.}}: Multi-stream  network trained without the procedure of label fusion. 
  \item \textbf{Multi-no-label}: Multi-stream densely connected network trained without the procedure of label fusion. 
  \item \textbf{DID-MDN (our)}: Multi-stream Densely-connected network trained with the procedure of estimated label fusion.
\end{itemize}

 \begin{figure*}[t]
 	\centering
 	\begin{minipage}{.162\textwidth}
 		\centering
		\caption*{\emph{PSNR: 16.47\\SSIM: 0.51}}
		\vskip-10pt
 		\includegraphics[width=2.8cm,height=2cm]{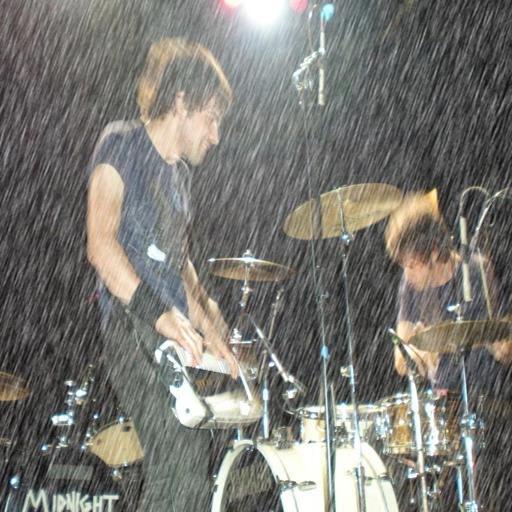}
 		\captionsetup{labelformat=empty}
 		\captionsetup{justification=centering}
		\vskip-2pt
		\caption*{\emph{Input}}
 	\end{minipage}
 	\begin{minipage}{.162\textwidth}
 		\centering
    	\caption*{\emph{PSNR: 22.87 \\SSIM: 0.8215}}
    	\vskip-10pt
 		\includegraphics[width=2.8cm,height=2cm]{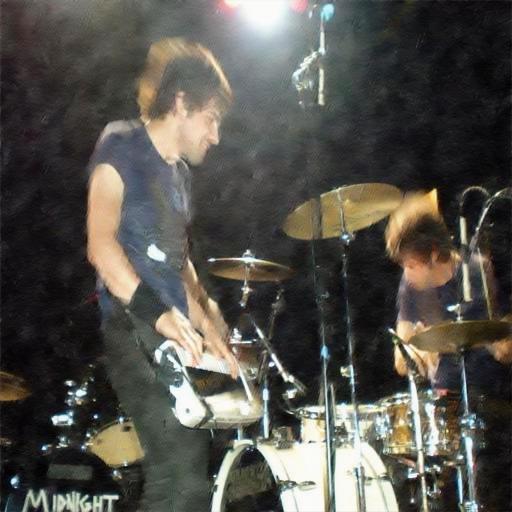}
 		\captionsetup{labelformat=empty}
 		\captionsetup{justification=centering}
		\vskip-2pt
		\caption*{\emph{Single}}
 	\end{minipage}
 	\begin{minipage}{.162\textwidth}
 		\centering
    	\caption*{\emph{PSNR: 23.02 \\SSIM: 0.8213}}
    	\vskip-10pt
 		\includegraphics[width=2.8cm,height=2cm]{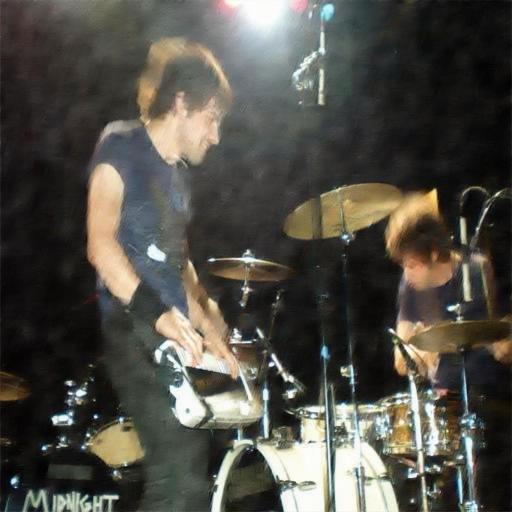}
 		\captionsetup{labelformat=empty}
 		\captionsetup{justification=centering}
		\vskip-2pt
		\caption*{\emph{Yang-Multi \cite{derain_cvpr2017_multi}}}
 	\end{minipage}
 	\begin{minipage}{.162\textwidth}
 		\centering
   		\caption*{\emph{PSNR: 23.47 \\SSIM: 0.8233}}
   		\vskip-10pt
 		\includegraphics[width=2.8cm,height=2cm]{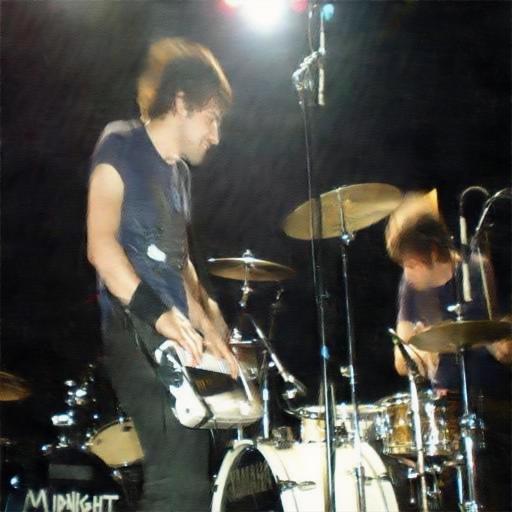}
 		\captionsetup{labelformat=empty}
 		\captionsetup{justification=centering}
		\vskip-2pt
		\caption*{\emph{Multi-no-label}}
 	\end{minipage}
 	\begin{minipage}{.16\textwidth}
 		\centering
  		\caption*{\emph{PSNR: \textbf{24.88} \\SSIM: \textbf{0.8623}}}
  		\vskip-10pt
 		\includegraphics[width=2.8cm,height=2cm]{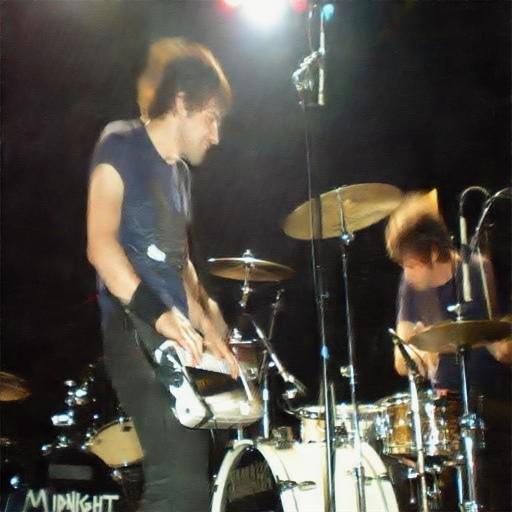}
 		\captionsetup{labelformat=empty}
 		\captionsetup{justification=centering}
		\vskip-2pt
		\caption*{\emph{DID-MDN}}
 	\end{minipage}
 	\begin{minipage}{.162\textwidth}
 		\centering
 		\caption*{\emph{PSNR: Inf \\SSIM: 1}}
 		\vskip-10pt
 		\includegraphics[width=2.8cm,height=2cm]{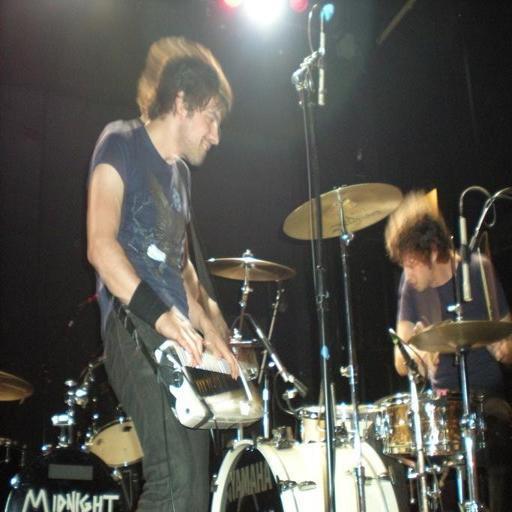}
 		\captionsetup{labelformat=empty}
 		\captionsetup{justification=centering}
		\vskip-2pt
		\caption*{\emph{Ground Truth}}
 	\end{minipage}
	\vskip-8pt
 	\caption{Results of ablation study on a synthetic image.} \label{fig:ablation}
 \end{figure*}

\begin{figure*}[htp!]
	\centering
	\begin{minipage}{.16\textwidth}
		\centering
		\caption*{\emph{PSNR: 17.27\\SSIM: 0.8257}}
		\vskip-10pt
 		\includegraphics[width=2.8cm,height=2cm]{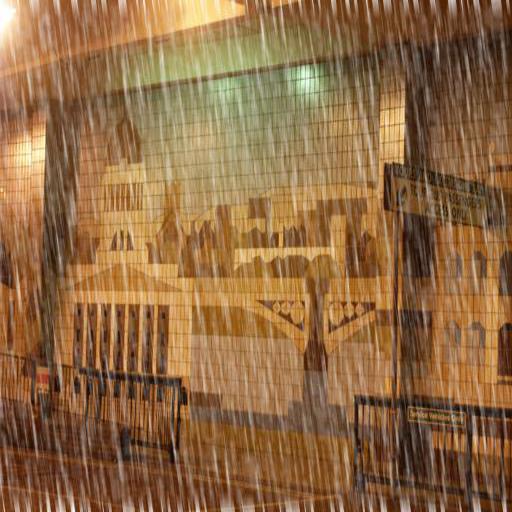}
		\captionsetup{labelformat=empty}
		\captionsetup{justification=centering}
	\end{minipage}  
	\begin{minipage}{.16\textwidth}
		\centering
		\caption*{\emph{PSNR:21.89 \\SSIM: 0.9007}}
		\vskip-10pt
 		\includegraphics[width=2.8cm,height=2cm]{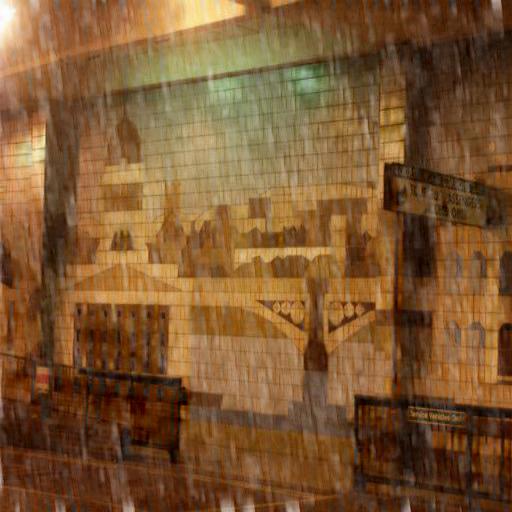}
		\captionsetup{labelformat=empty}
		\captionsetup{justification=centering}
	\end{minipage}
	\begin{minipage}{.16\textwidth}
		\centering
		\caption*{\emph{PSNR: 25.30 \\SSIM:0.9455}}
		\vskip-10pt
 		\includegraphics[width=2.8cm,height=2cm]{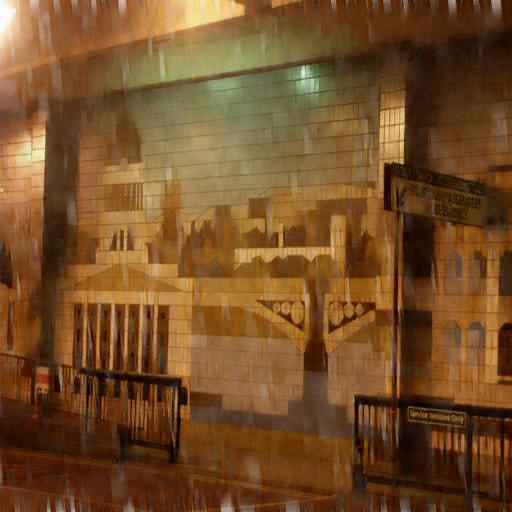}
		\captionsetup{labelformat=empty}
		\captionsetup{justification=centering}
	\end{minipage}
	\begin{minipage}{.16\textwidth}
		\centering
		\caption*{\emph{PSNR: 20.72 \\SSIM: 0.8885}}
		\vskip-10pt
 		\includegraphics[width=2.8cm,height=2cm]{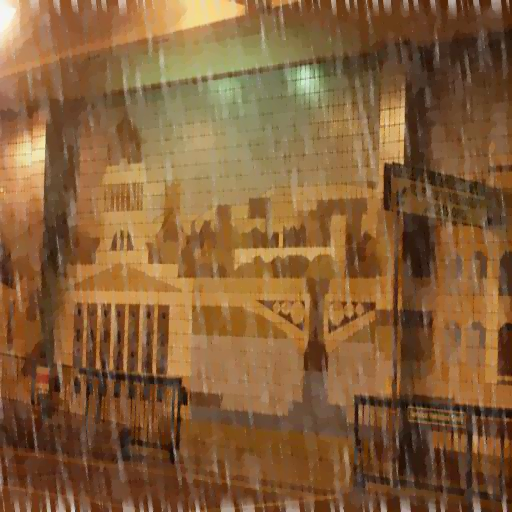}
		\captionsetup{labelformat=empty}
		\captionsetup{justification=centering}
	\end{minipage}
	\begin{minipage}{.16\textwidth}
		\centering
		\caption*{\emph{PSNR: \textbf{25.95} \\SSIM: \textbf{0.9605}}}
		\vskip-10pt
 		\includegraphics[width=2.8cm,height=2cm]{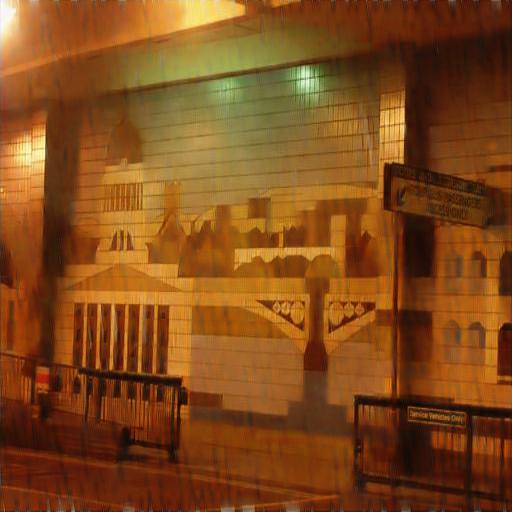}
		\captionsetup{labelformat=empty}
		\captionsetup{justification=centering}
	\end{minipage} 
	\begin{minipage}{.16\textwidth}
		\centering
		\caption*{\emph{PSNR: Inf\\SSIM: 1}}
		\vskip-10pt
 		\includegraphics[width=2.8cm,height=2cm]{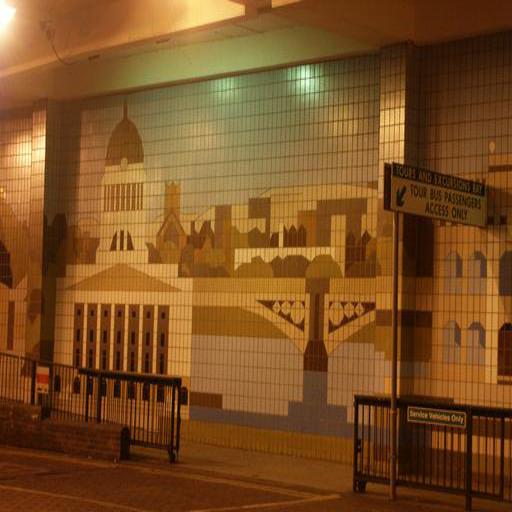}
		\captionsetup{labelformat=empty}
		\captionsetup{justification=centering}
	\end{minipage}\\		\vskip+6pt		
	\begin{minipage}{.16\textwidth}
		\centering
		\caption*{\emph{PSNR:19.31 \\SSIM: 0.7256}}
		\vskip-10pt
 		\includegraphics[width=2.8cm,height=2cm]{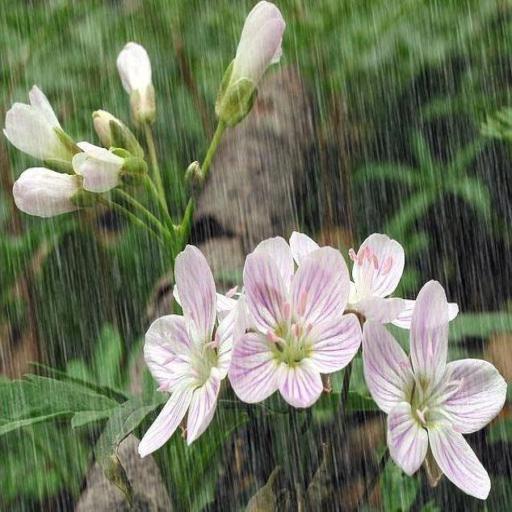}
		\captionsetup{labelformat=empty}
		\captionsetup{justification=centering}
	\end{minipage} 
	\begin{minipage}{.16\textwidth}
		\centering
		\caption*{\emph{PSNR:22.28 \\SSIM: 0.8199}}
		\vskip-10pt
 		\includegraphics[width=2.8cm,height=2cm]{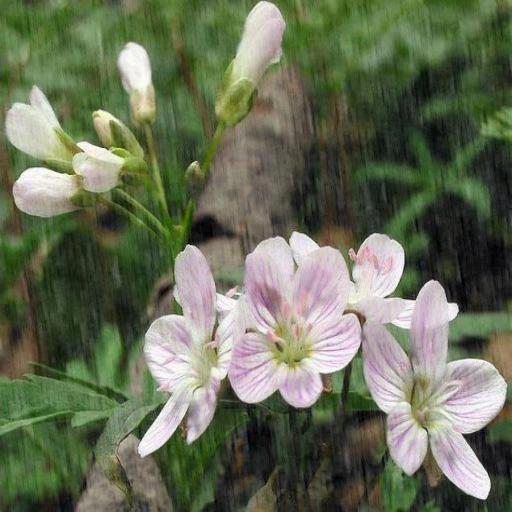}
		\captionsetup{labelformat=empty}
		\captionsetup{justification=centering}

	\end{minipage} 
	\begin{minipage}{.16\textwidth}
		\centering
		\caption*{\emph{PSNR:26.88\\SSIM:0.8814}}
		\vskip-10pt
 		\includegraphics[width=2.8cm,height=2cm]{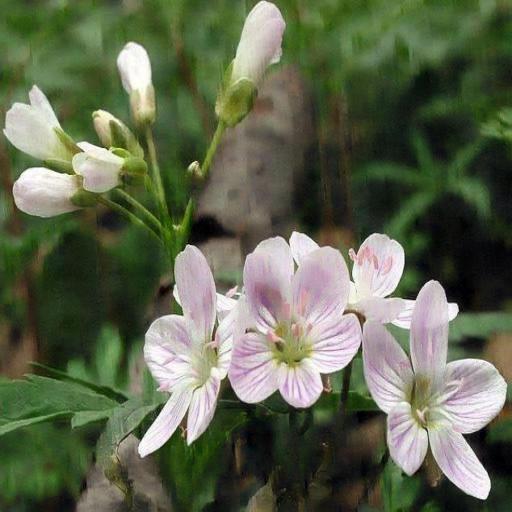}
		\captionsetup{labelformat=empty}
		\captionsetup{justification=centering}
	\end{minipage} 
	\begin{minipage}{.16\textwidth}
		\centering
		\caption*{\emph{PSNR: 21.42 \\SSIM:0.7878}}
		\vskip-10pt
 		\includegraphics[width=2.8cm,height=2cm]{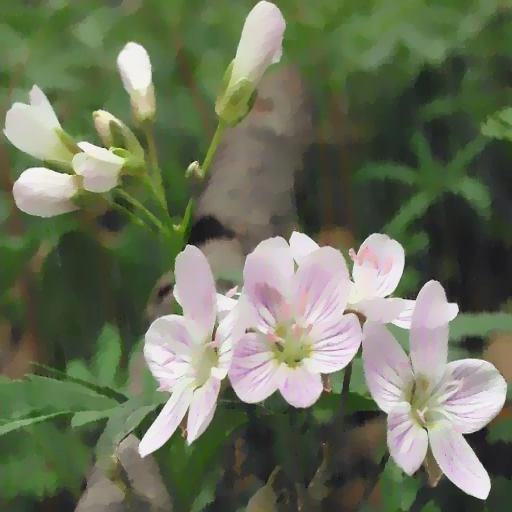}
		\captionsetup{labelformat=empty}
		\captionsetup{justification=centering}
	\end{minipage} 
	\begin{minipage}{.16\textwidth}
		\centering
		\caption*{\emph{PSNR: \textbf{29.88}\\SSIM:\textbf{0.9252}}}
		\vskip-10pt
 		\includegraphics[width=2.8cm,height=2cm]{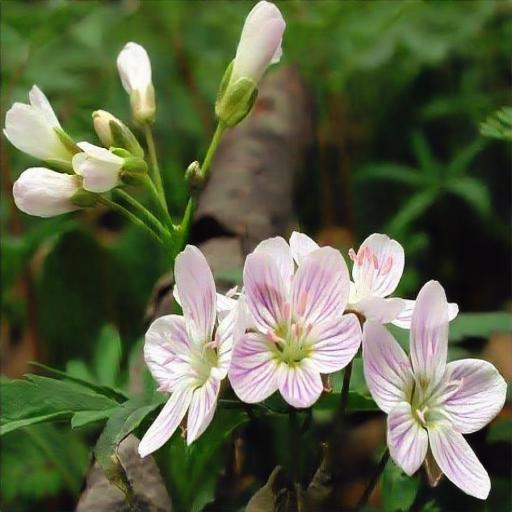}
		\captionsetup{labelformat=empty}
		\captionsetup{justification=centering}
	\end{minipage}
	\begin{minipage}{.16\textwidth}
		\centering
		\caption*{\emph{PSNR: Inf \\SSIM:1}}
		\vskip-10pt
 		\includegraphics[width=2.8cm,height=2cm]{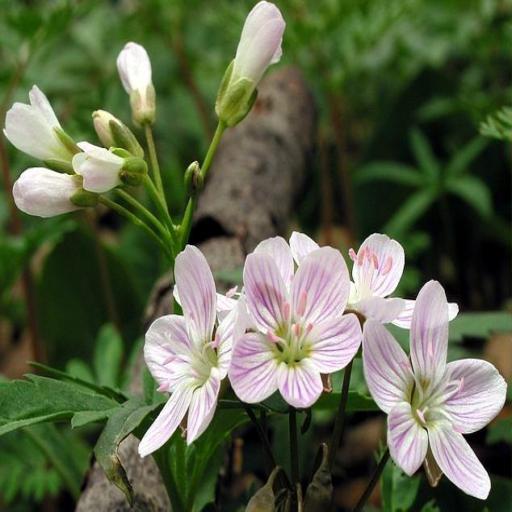}
		\captionsetup{labelformat=empty}
		\captionsetup{justification=centering}
	\end{minipage} \\	\vskip+6pt
	\begin{minipage}{.16\textwidth}
		\centering
		\caption*{\emph{PSNR: 20.74\\SSIM:0.7992}}
		\vskip-10pt
 		\includegraphics[width=2.8cm,height=2.8cm]{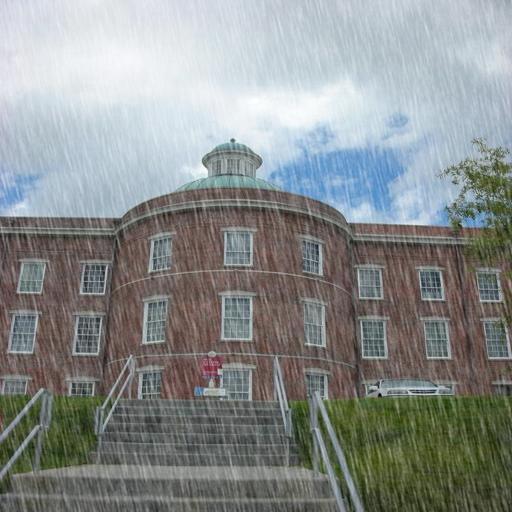}
		\captionsetup{labelformat=empty}
		\captionsetup{justification=centering}
		\caption*{\emph{Input} \\ \quad}
	\end{minipage} 
	\begin{minipage}{.16\textwidth}
		\centering
		\caption*{\emph{PSNR:24.20\\SSIM:0.8502}}
		\vskip-10pt
 		\includegraphics[width=2.8cm,height=2.8cm]{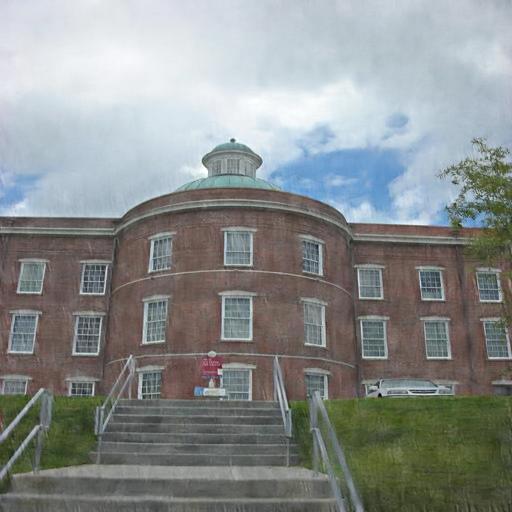}
		\captionsetup{labelformat=empty}
		\captionsetup{justification=centering}
		\caption*{\emph{JORDER (CVPR'17) \cite{derain_cvpr2017_multi}}}
	\end{minipage} 
	\begin{minipage}{.16\textwidth}
		\centering
		\caption*{\emph{PSNR:29.44\\SSIM:0.9429}}
		\vskip-10pt
 		\includegraphics[width=2.8cm,height=2.8cm]{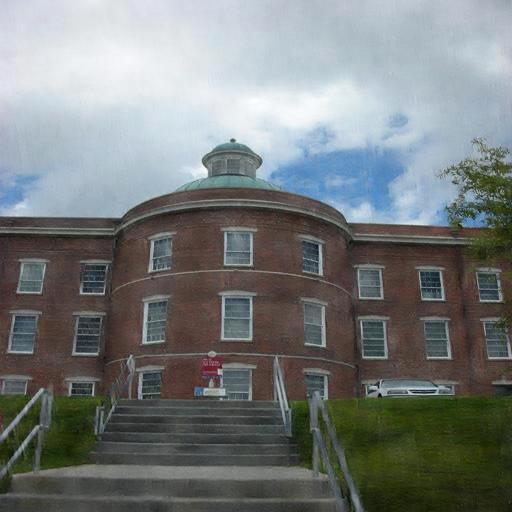}
		\captionsetup{labelformat=empty}
		\captionsetup{justification=centering}
		\caption*{\emph{DDN (CVPR'17) \\\cite{derain_cvpr2017}}}
	\end{minipage} 
	\begin{minipage}{.16\textwidth}
		\centering
		\caption*{\emph{PSNR:25.32\\ SSIM: 0.8922}}
		\vskip-10pt
 		\includegraphics[width=2.8cm,height=2.8cm]{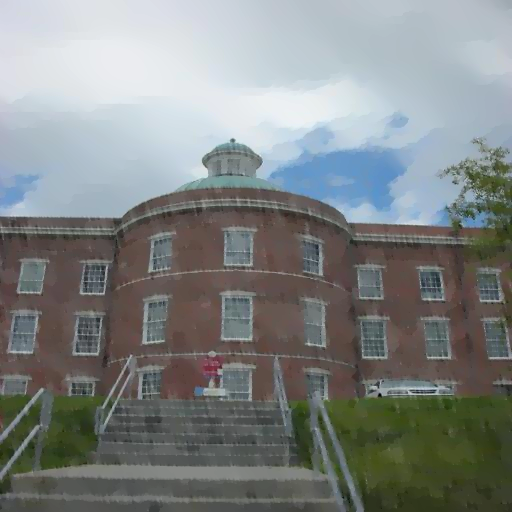}
		\captionsetup{labelformat=empty}
		\captionsetup{justification=centering}
		\caption*{\emph{JBO (ICCV'17) \\ \cite{derain_iccv17}}}
	\end{minipage} 
	\begin{minipage}{.16\textwidth}
		\centering
		\caption*{\emph{PSNR:\textbf{29.84}\\SSIM:\textbf{0.9482}}}
		\vskip-10pt
 		\includegraphics[width=2.8cm,height=2.8cm]{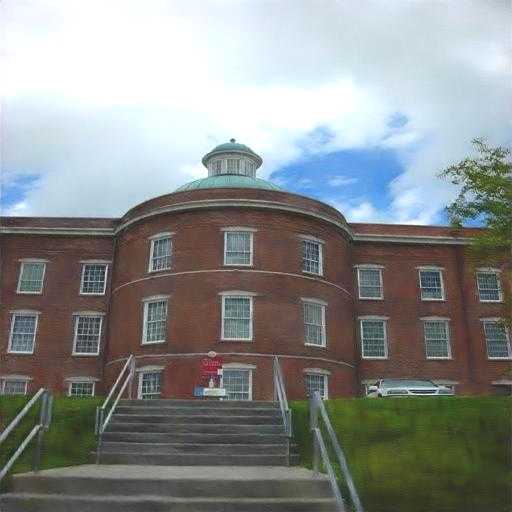}
		\captionsetup{labelformat=empty}
		\captionsetup{justification=centering}
		\caption*{\emph{DID-MDN} \\ \quad}
	\end{minipage}
	\begin{minipage}{.16\textwidth}
		\centering
		\caption*{\emph{PSNR: Inf \\SSIM:1}}
		\vskip-10pt
 		\includegraphics[width=2.8cm,height=2.8cm]{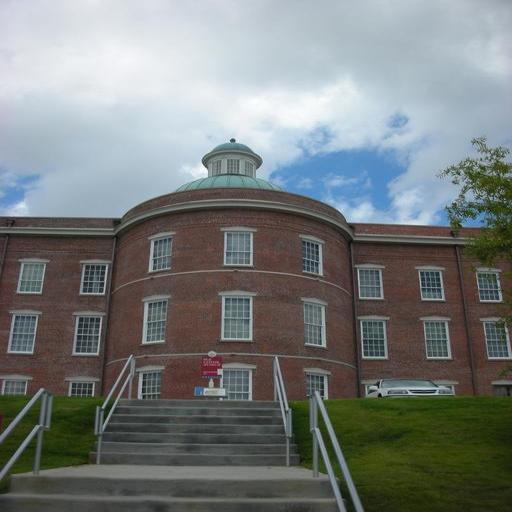}
		\captionsetup{labelformat=empty}
		\captionsetup{justification=centering}
		\caption*{\emph{Ground Truth} \\ \quad}
	\end{minipage}
	\vskip -10pt
	\caption{ Rain-streak removal results on sample images from the synthetic datasets \emph{Test1} and \emph{Test2}.}\label{syn_rainy}
\end{figure*}

The average PSNR and SSIM results evaluated on \emph{Test1} are tabulated in Table \ref{tab:baselinetable}.   As shown in Fig.~\ref{fig:ablation}, even though the single stream network and Yang's multi-stream network \cite{derain_cvpr2017_multi} are able to successfully remove the rain streak components, they both tend to over de-rain the image with the blurry output.  The multi-stream network without label fusion is unable to accurately  estimate the rain-density level and hence it tends to leave some rain streaks in the de-rained image (especially observed from the derained-part around the light).   In contrast, the proposed multi-stream network with label fusion approach is capable of removing rain streaks while preserving the background details. Similar observations can be made using the quantitative results as shown in Table~\ref{tab:baselinetable}.

\begin{figure*}[htp!]
	\centering
	\begin{minipage}{.195\textwidth}
		\centering
		\includegraphics[width=3.45cm,height=2.2cm]{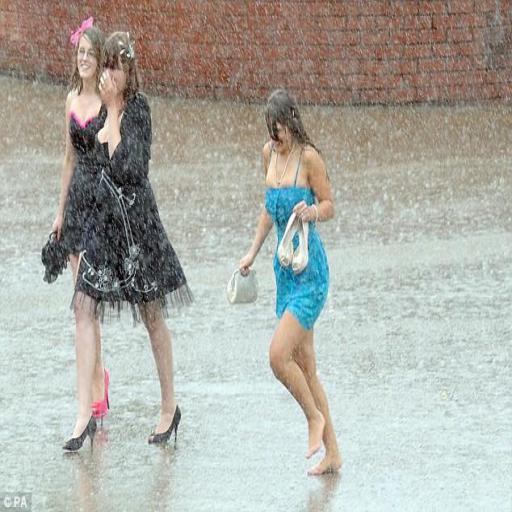}
		\captionsetup{labelformat=empty}
		\captionsetup{justification=centering}
	\end{minipage}  
	\begin{minipage}{.195\textwidth}
		\centering
		\includegraphics[width=3.45cm,height=2.2cm]{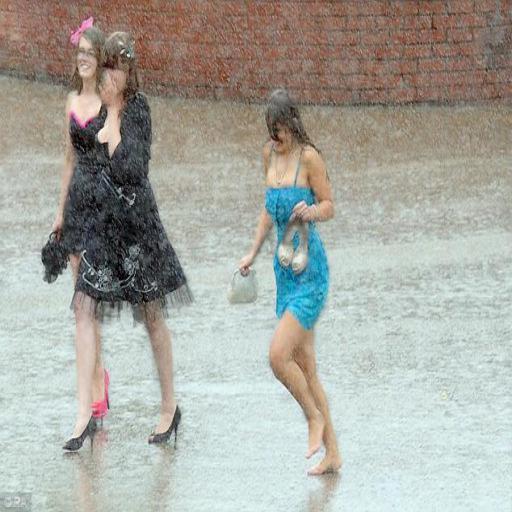}
		\captionsetup{labelformat=empty}
		\captionsetup{justification=centering}
	\end{minipage}
	\begin{minipage}{.195\textwidth}
		\centering
		\includegraphics[width=3.45cm,height=2.2cm]{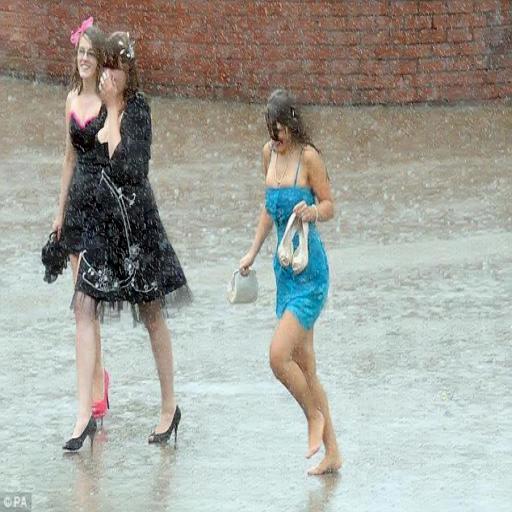}
		\captionsetup{labelformat=empty}
		\captionsetup{justification=centering}
	\end{minipage}
	\begin{minipage}{.195\textwidth}
		\centering
		\includegraphics[width=3.45cm,height=2.2cm]{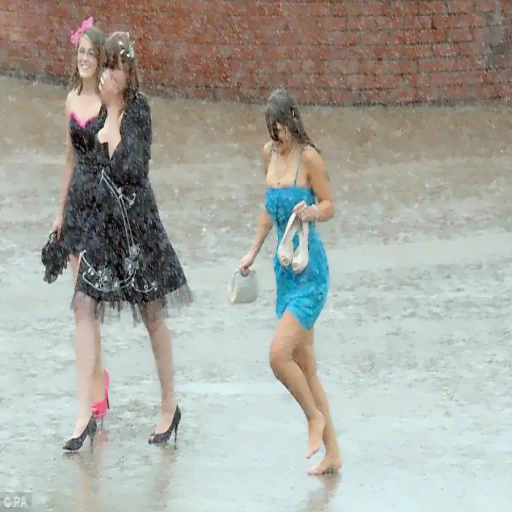}
		\captionsetup{labelformat=empty}
		\captionsetup{justification=centering}
	\end{minipage}
	\begin{minipage}{.195\textwidth}
		\centering
		\includegraphics[width=3.45cm,height=2.2cm]{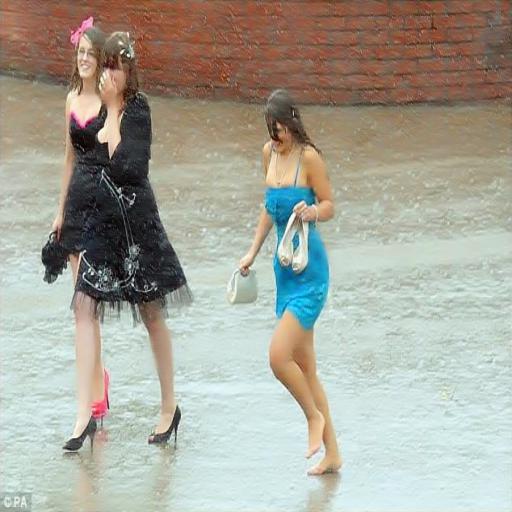}
		\captionsetup{labelformat=empty}
		\captionsetup{justification=centering}
	\end{minipage} \\		\vskip+1pt
	\begin{minipage}{.195\textwidth}
		\centering
		\includegraphics[width=3.45cm,height=2.2cm]{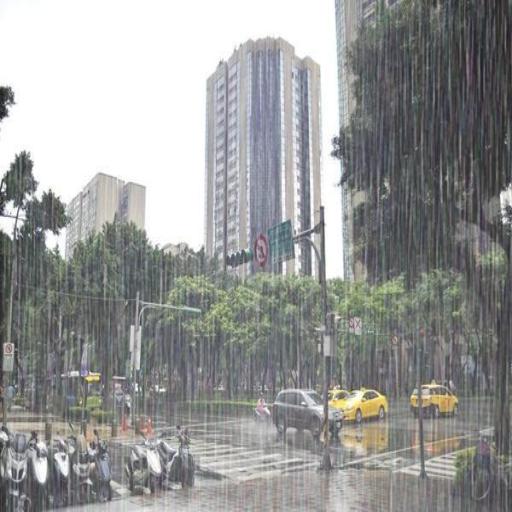}
		\captionsetup{labelformat=empty}
		\captionsetup{justification=centering}
	\end{minipage}  
	\begin{minipage}{.195\textwidth}
		\centering
		\includegraphics[width=3.45cm,height=2.2cm]{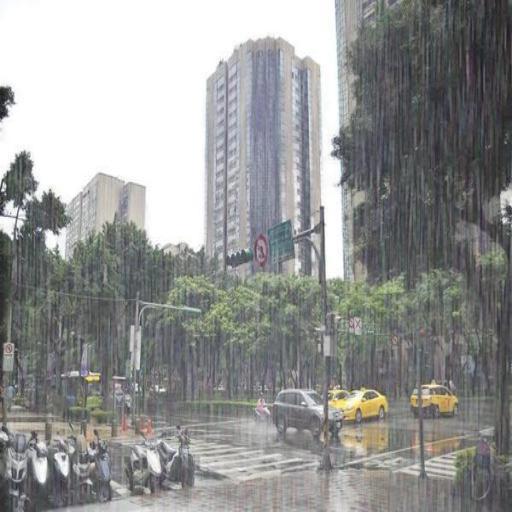}
		\captionsetup{labelformat=empty}
		\captionsetup{justification=centering}
	\end{minipage}
	\begin{minipage}{.195\textwidth}
		\centering
		\includegraphics[width=3.45cm,height=2.2cm]{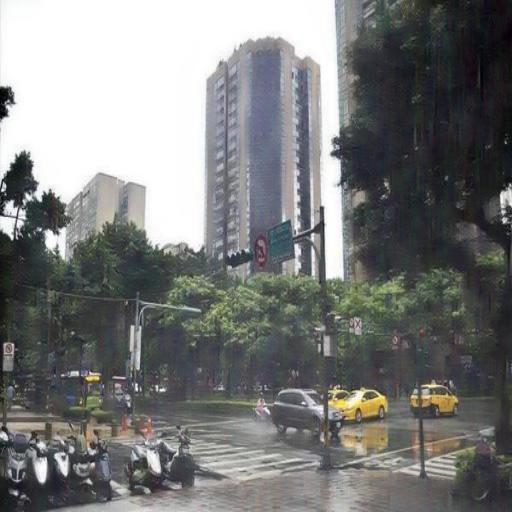}
		\captionsetup{labelformat=empty}
		\captionsetup{justification=centering}
	\end{minipage}
	\begin{minipage}{.195\textwidth}
		\centering
		\includegraphics[width=3.45cm,height=2.2cm]{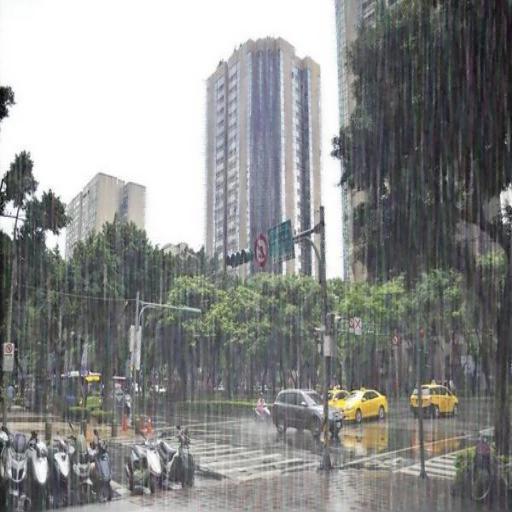}
		\captionsetup{labelformat=empty}
		\captionsetup{justification=centering}
	\end{minipage}
	\begin{minipage}{.195\textwidth}
		\centering
		\includegraphics[width=3.45cm,height=2.2cm]{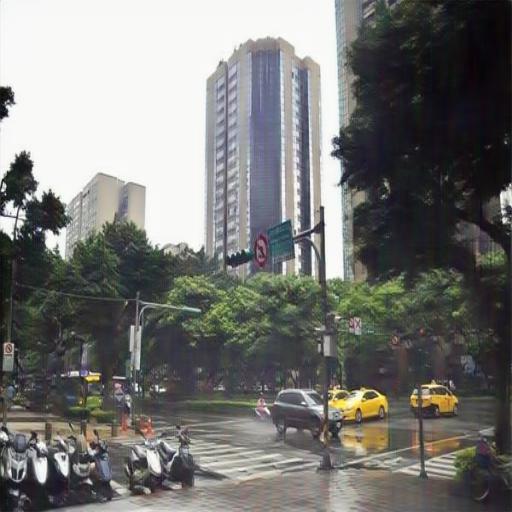}
		\captionsetup{labelformat=empty}
		\captionsetup{justification=centering}
	\end{minipage} \\		\vskip+1pt
	\begin{minipage}{.195\textwidth}
		\centering
		\includegraphics[width=3.45cm,height=2.4cm]{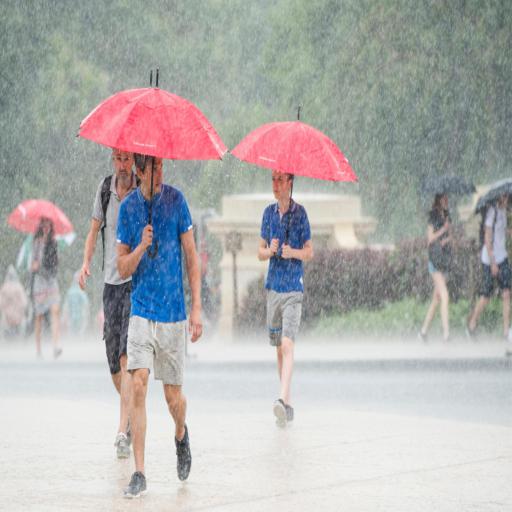}
		\captionsetup{labelformat=empty}
		\captionsetup{justification=centering}
	\end{minipage}  
	\begin{minipage}{.195\textwidth}
		\centering
		\includegraphics[width=3.45cm,height=2.4cm]{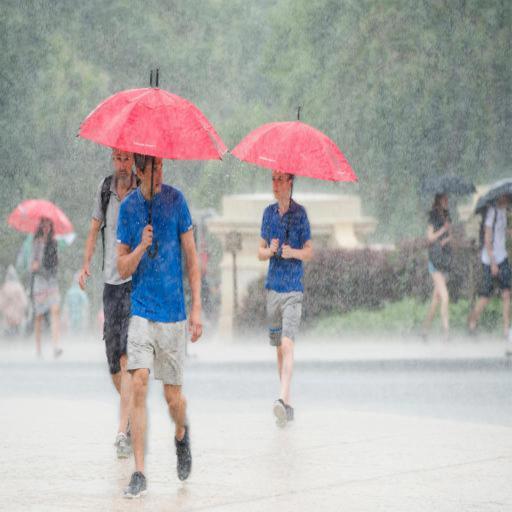}
		\captionsetup{labelformat=empty}
		\captionsetup{justification=centering}
	\end{minipage}
	\begin{minipage}{.195\textwidth}
		\centering
		\includegraphics[width=3.45cm,height=2.4cm]{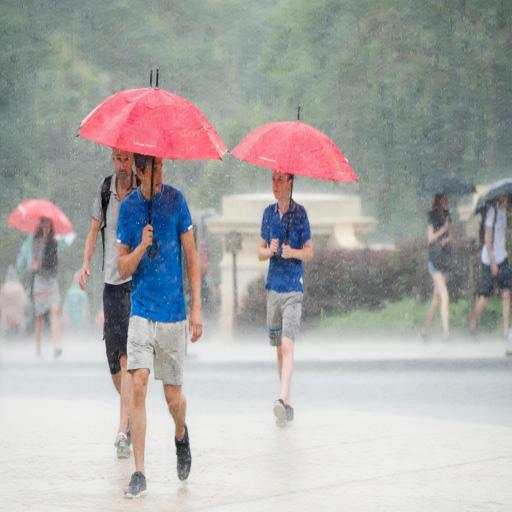}
		\captionsetup{labelformat=empty}
		\captionsetup{justification=centering}
	\end{minipage}
	\begin{minipage}{.195\textwidth}
		\centering
		\includegraphics[width=3.45cm,height=2.4cm]{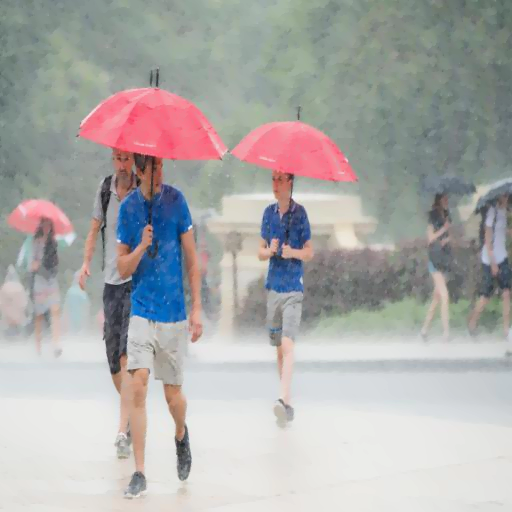}
		\captionsetup{labelformat=empty}
		\captionsetup{justification=centering}
	\end{minipage}
	\begin{minipage}{.195\textwidth}
		\centering
		\includegraphics[width=3.45cm,height=2.4cm]{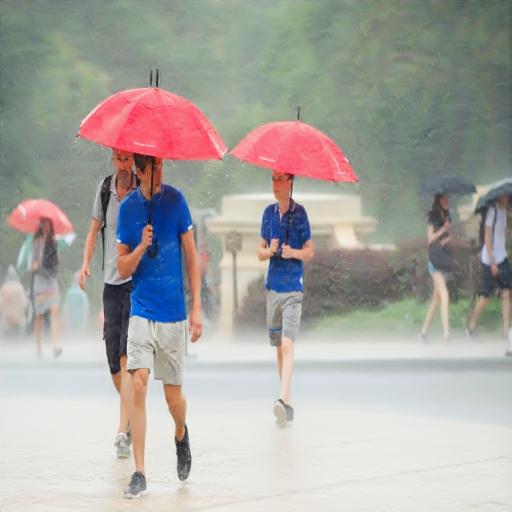}
		\captionsetup{labelformat=empty}
		\captionsetup{justification=centering}
	\end{minipage} \\		\vskip+1pt
	\begin{minipage}{.195\textwidth}
		\centering
		\includegraphics[width=3.45cm,height=2.4cm]{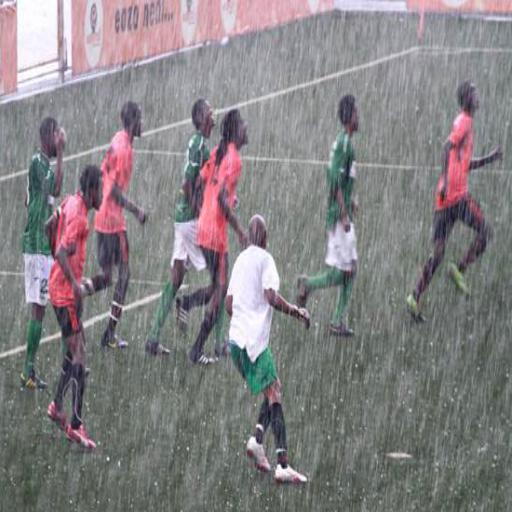}
		\captionsetup{labelformat=empty}
		\captionsetup{justification=centering}
	\end{minipage}  
	\begin{minipage}{.195\textwidth}
		\centering
		\includegraphics[width=3.45cm,height=2.4cm]{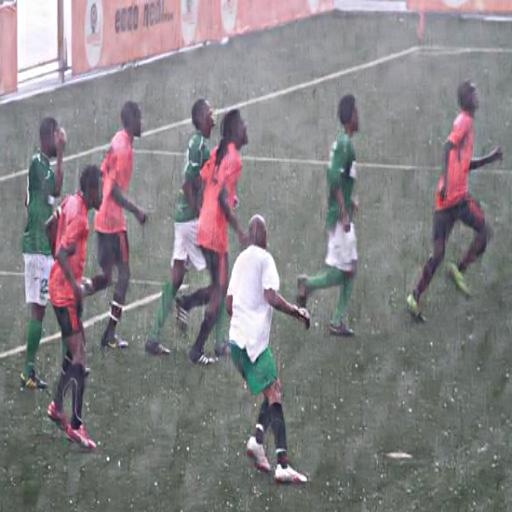}
		\captionsetup{labelformat=empty}
		\captionsetup{justification=centering}
	\end{minipage}
	\begin{minipage}{.195\textwidth}
		\centering
		\includegraphics[width=3.45cm,height=2.4cm]{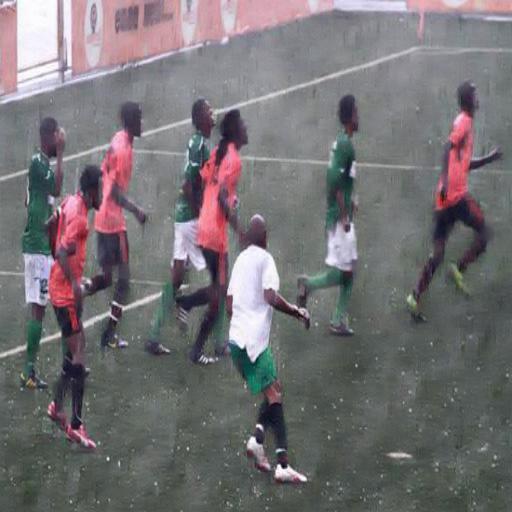}
		\captionsetup{labelformat=empty}
		\captionsetup{justification=centering}
	\end{minipage}
	\begin{minipage}{.195\textwidth}
		\centering
		\includegraphics[width=3.45cm,height=2.4cm]{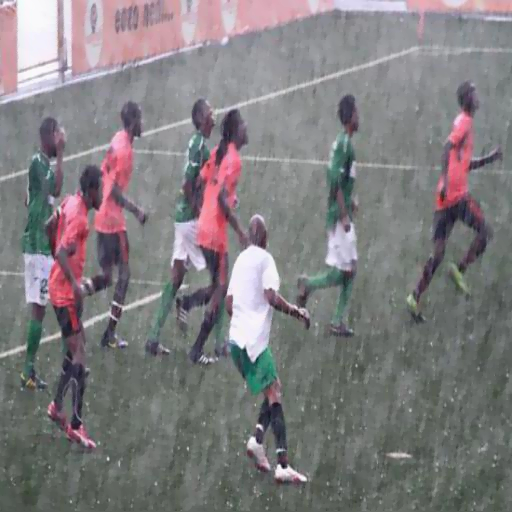}
		\captionsetup{labelformat=empty}
		\captionsetup{justification=centering}
	\end{minipage}
	\begin{minipage}{.195\textwidth}
		\centering
		\includegraphics[width=3.45cm,height=2.4cm]{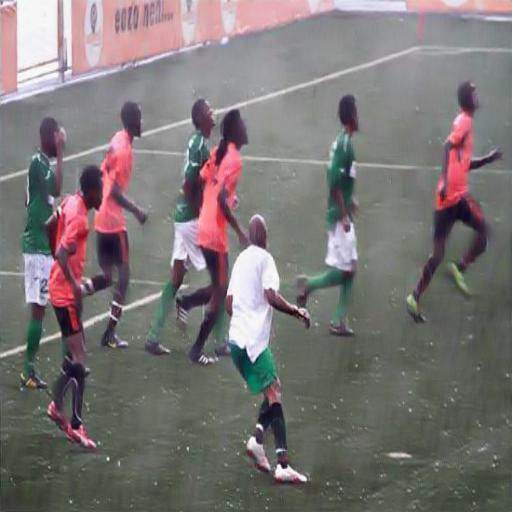}
		\captionsetup{labelformat=empty}
		\captionsetup{justification=centering}
	\end{minipage} \\		\vskip+1pt
	\begin{minipage}{.195\textwidth}
		\centering
		\includegraphics[width=3.45cm,height=2.25cm]{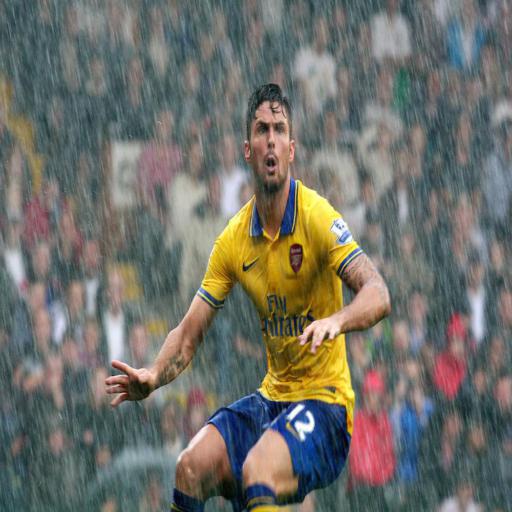}
		\captionsetup{labelformat=empty}
		\captionsetup{justification=centering}
		\caption*{\emph{Input} \\ \quad}
	\end{minipage}  
	\begin{minipage}{.195\textwidth}
		\centering
		\includegraphics[width=3.45cm,height=2.25cm]{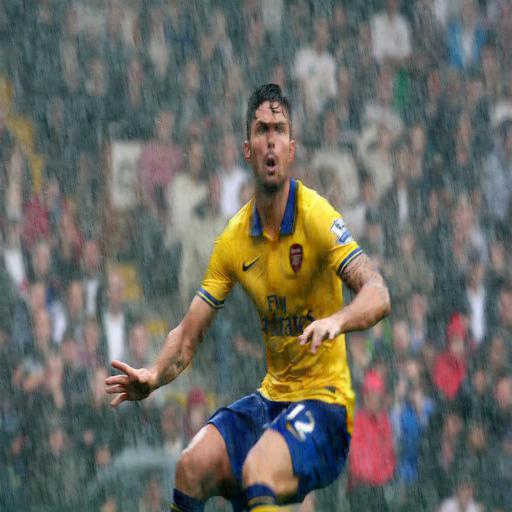}
		\captionsetup{labelformat=empty}
		\captionsetup{justification=centering}
		\caption*{\emph{JORDER (CVPR'17)  \\ \cite{derain_cvpr2017_multi}}}
	\end{minipage}
	\begin{minipage}{.195\textwidth}
		\centering
		\includegraphics[width=3.45cm,height=2.25cm]{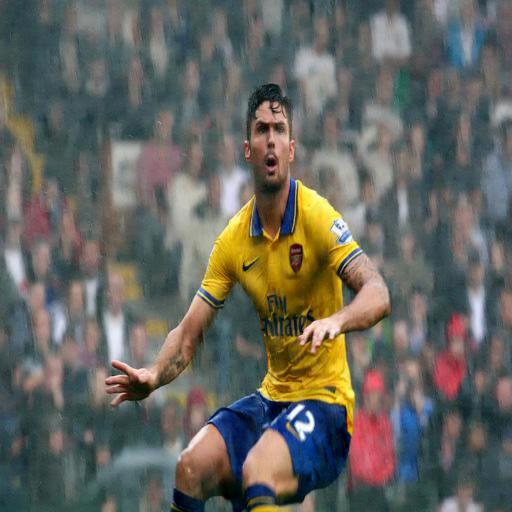}
		\captionsetup{labelformat=empty}
		\captionsetup{justification=centering}
		\caption*{\emph{DDN (CVPR'17) \\ \cite{derain_cvpr2017}}}
	\end{minipage}
	\begin{minipage}{.195\textwidth}
		\centering
		\includegraphics[width=3.45cm,height=2.25cm]{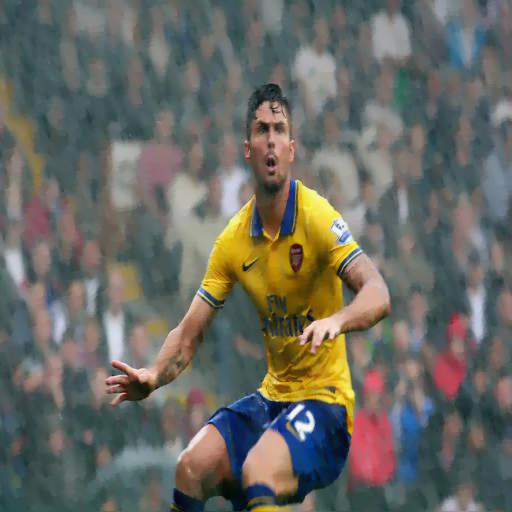}
		\captionsetup{labelformat=empty}
		\captionsetup{justification=centering}
		\caption*{\emph{JBO (ICCV'17) \\ \cite{derain_iccv17}}}
	\end{minipage}
	\begin{minipage}{.195\textwidth}
		\centering
		\includegraphics[width=3.45cm,height=2.25cm]{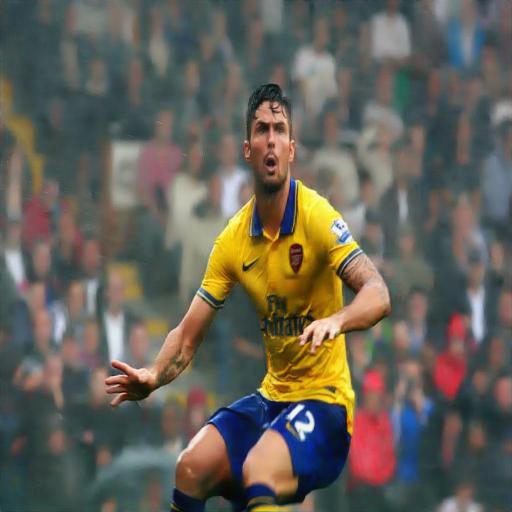}
		\captionsetup{labelformat=empty}
		\captionsetup{justification=centering}
		\caption*{\emph{DID-MDN} \\ \qquad}
	\end{minipage} 
	\caption{Rain-streak removal results on sample real-world images.}\label{reaL-rainy2}
\end{figure*}

\subsubsection{Results on Two Synthetic Datasets}
We compare quantitative and qualitative performance of different methods on the test images from the two synthetic datasets - \emph{Test1} and \emph{Test2}. Quantitative results corresponding to different methods are tabulated in Table~\ref{ta:quantitive}. It can be clearly observed that the proposed DID-MDN is able to achieve superior quantitative performance.  

To visually demonstrate the improvements obtained by the proposed method on the synthetic dataset, results on two sample images selected from \emph{Test2} and one sample chosen from our newly synthesized \emph{Test1} are presented in Figure~\ref{syn_rainy}. Note that we selectively sample  images from all three conditions to show that our method performs well under different variations \footnote{Due to space limitations and for better comparisons, we only show the results corresponding to the most recent state-of-the-art methods \cite{derain_cvpr2017_multi,derain_cvpr2017,derain_iccv17} in the main paper.  More results corresponding to the other methods \cite{dis_rain_2015,rain_2016_gmm,derain_tip17} can be found in  \emph{Supplementary Material}.}. While the JORDER method \cite{derain_cvpr2017_multi} is able to remove some parts of the rain-streaks, it still tends  to leave some rain-streaks in the de-rained images. Similar results are also observed from \cite{derain_iccv17}. Even though the method of Fu~\emph{et al.} \cite{derain_cvpr2017} is able to remove the rain-streak, especially in the medium and light rain conditions, it tends to remove some important details as well, such as  flower details, as shown in the second row and window structures as shown in the third row (Details can be better observed via zooming-in the figure). Overall, the proposed method is able to preserve better details while effectively removing the rain-streak components. 

\subsubsection{Results on Real-World Images}
The  performance  of  the
proposed  method is also evaluated on  many  real-world images downloaded from the Internet and also real-world images published by the authors of \cite{derain_2017_zhang,derain_cvpr2017}.  The de-raining results are shown in Fig~\ref{reaL-rainy2}.

As before, previous methods either tend to under de-rain or over de-rain the images. In contrast, the proposed method achieves better results in terms of effectively removing rain streaks while preserving the image details.  In addition,   it can be observed that the proposed method is able to deal with different types of rain conditions, such as heavy rain shown in the second row of Fig~\ref{reaL-rainy2} and medium rain shown in the fifth row of Fig~\ref{reaL-rainy2}.  Furthermore, the proposed method can effectively deal with rain-streaks containing different shapes and scales such as small round rain streaks shown in the third row in Fig~\ref{reaL-rainy2} and long-thin rain-streak in the second row in Fig~\ref{reaL-rainy2}. Overall, the results evaluated on real-world images captured from different rain conditions demonstrate the effectiveness and the robustness of the proposed \emph{DID-MDN} method. More results can be found in \emph{Supplementary Material}.

\subsubsection{Running Time Comparisons}
Running time comparisons are shown in the table below.   It can be observed that the testing time of the proposed DID-MDN is comparable to the DDN \cite{derain_cvpr2017} method.  On average, it takes about 0.3s to de-rain an image of size $512\times 512$. 
\begin{table}[ht!]
	\centering
	\caption{Running time (in seconds) for different methods averaged on 1000 images  with size 512$\times$512. }
	\label{ta:time_eff}
	\resizebox{0.5\textwidth}{!}{%
		\begin{tabular}{|c|c|c|c|c|c|c|c|}
			\hline
			& DSC  & GMM & CNN (GPU) & JORDER (GPU) & DDN (GPU) & JBO (CPU)  & DID-MDN (GPU) \\ \hline\hline
			512X512 & 189.3s & 674.8s  & 2.8s  & 600.6s & 0.3s & 1.4s & \textbf{0.2s} \\ \hline
		\end{tabular}}
	\end{table}

\section{Conclusion}
In this paper, we propose a novel density-aware image deraining method with multi-stream densely connected network (DID-MDN) for jointly rain-density estimation and deraining. In comparison to existing approaches which attempt to solve the de-raining problem using a single network to learn to remove rain streaks with different densities (heavy, medium and light), we investigated the use of estimated rain-density label for guiding the synthesis of the de-rained image. To efficiently predict the rain-density label, a residual-aware rain-density classier is proposed in this paper.   Detailed experiments and comparisons are performed on two synthetic and one real-world datasets to demonstrate that the proposed DID-MDN method significantly outperforms many recent state-of-the-art methods.
Additionally, the proposed DID-MDN method is compared against baseline configurations to illustrate the performance gains obtained by each module.

{\small
\bibliographystyle{ieee}
\bibliography{egbib}

\begin{thebibliography}{10}\itemsep=-1pt

\bibitem{derain_depth_sparse}
D.-Y. Chen, C.-C. Chen, and L.-W. Kang.
\newblock Visual depth guided color image rain streaks removal using sparse
  coding.
\newblock {\em IEEE transactions on circuits and systems for video technology},
  24(8):1430--1455, 2014.

\bibitem{derain_lowrank}
Y.-L. Chen and C.-T. Hsu.
\newblock A generalized low-rank appearance model for spatio-temporally
  correlated rain streaks.
\newblock In {\em IEEE ICCV}, pages 1968--1975, 2013.

\bibitem{cnn_derain2}
D.~Eigen, D.~Krishnan, and R.~Fergus.
\newblock Restoring an image taken through a window covered with dirt or rain.
\newblock In {\em ICCV}, pages 633--640, 2013.

\bibitem{depth_nips_14}
D.~Eigen, C.~Puhrsch, and R.~Fergus.
\newblock Depth map prediction from a single image using a multi-scale deep
  network.
\newblock In {\em NIPS}, pages 2366--2374, 2014.

\bibitem{derain_tip17}
X.~Fu, J.~Huang, X.~Ding, Y.~Liao, and J.~Paisley.
\newblock Clearing the skies: A deep network architecture for single-image rain
  removal.
\newblock {\em IEEE Transactions on Image Processing}, 26(6):2944--2956, 2017.

\bibitem{derain_cvpr2017}
X.~Fu, J.~Huang, D.~Zeng, Y.~Huang, X.~Ding, and J.~Paisley.
\newblock Removing rain from single images via a deep detail network.
\newblock In {\em 2017 IEEE Conference on Computer Vision and Pattern
  Recognition (CVPR)}, pages 1715--1723, July 2017.

\bibitem{spp}
K.~He, X.~Zhang, S.~Ren, and J.~Sun.
\newblock Spatial pyramid pooling in deep convolutional networks for visual
  recognition.
\newblock In {\em European Conference on Computer Vision}, pages 346--361.
  Springer, 2014.

\bibitem{deep_residue}
K.~He, X.~Zhang, S.~Ren, and J.~Sun.
\newblock Deep residual learning for image recognition.
\newblock In {\em Proceedings of the IEEE Conference on Computer Vision and
  Pattern Recognition}, pages 770--778, 2016.

\bibitem{derain_tip14}
D.-A. Huang, L.-W. Kang, Y.-C.~F. Wang, and C.-W. Lin.
\newblock Self-learning based image decomposition with applications to single
  image denoising.
\newblock {\em IEEE Transactions on multimedia}, 16(1):83--93, 2014.

\bibitem{derain_context}
D.-A. Huang, L.-W. Kang, M.-C. Yang, C.-W. Lin, and Y.-C.~F. Wang.
\newblock Context-aware single image rain removal.
\newblock In {\em Multimedia and Expo (ICME), 2012 IEEE International
  Conference on}, pages 164--169. IEEE, 2012.

\bibitem{dense_net}
G.~Huang, Z.~Liu, K.~Q. Weinberger, and L.~van~der Maaten.
\newblock Densely connected convolutional networks.
\newblock {\em arXiv preprint arXiv:1608.06993}, 2016.

\bibitem{dense_fully}
S.~J{\'e}gou, M.~Drozdzal, D.~Vazquez, A.~Romero, and Y.~Bengio.
\newblock The one hundred layers tiramisu: Fully convolutional densenets for
  semantic segmentation.
\newblock In {\em Computer Vision and Pattern Recognition Workshops (CVPRW),
  2017 IEEE Conference on}, pages 1175--1183. IEEE, 2017.

\bibitem{perceptual_loss}
J.~Johnson, A.~Alahi, and L.~Fei-Fei.
\newblock Perceptual losses for real-time style transfer and super-resolution.
\newblock In {\em European Conference on Computer Vision}, pages 694--711.
  Springer, 2016.

\bibitem{derain_tip12}
L.-W. Kang, C.-W. Lin, and Y.-H. Fu.
\newblock Automatic single-image-based rain streaks removal via image
  decomposition.
\newblock {\em IEEE TIP}, 21(4):1742--1755, 2012.

\bibitem{SR_photorea}
C.~Ledig, L.~Theis, F.~Husz{\'a}r, J.~Caballero, A.~Cunningham, A.~Acosta,
  A.~Aitken, A.~Tejani, J.~Totz, Z.~Wang, et~al.
\newblock Photo-realistic single image super-resolution using a generative
  adversarial network.
\newblock In {\em Proceedings of the IEEE Conference on Computer Vision and
  Pattern Recognition}, pages 1--8, 2017.

\bibitem{kunpeng_ijcai}
K.~Li, Y.~Kong, and Y.~Fu.
\newblock Multi-stream deep similarity learning networks for visual tracking.
\newblock In {\em IJCAI}, 2017.

\bibitem{rain_2016_gmm}
Y.~Li, R.~T. Tan, X.~Guo, J.~Lu, and M.~S. Brown.
\newblock Rain streak removal using layer priors.
\newblock In {\em 2016 IEEE Conference on Computer Vision and Pattern
  Recognition (CVPR)}, pages 2736--2744, June 2016.

\bibitem{fcn}
J.~Long, E.~Shelhamer, and T.~Darrell.
\newblock Fully convolutional networks for semantic segmentation.
\newblock In {\em Proceedings of the IEEE Conference on Computer Vision and
  Pattern Recognition}, pages 3431--3440, 2015.

\bibitem{dis_rain_2015}
Y.~Luo, Y.~Xu, and H.~Ji.
\newblock Removing rain from a single image via discriminative sparse coding.
\newblock In {\em ICCV}, pages 3397--3405, 2015.

\bibitem{peng_face}
X.~Peng, R.~S. Feris, X.~Wang, and D.~N. Metaxas.
\newblock A recurrent encoder-decoder network for sequential face alignment.
\newblock In {\em European Conference on Computer Vision}, pages 38--56.
  Springer International Publishing, 2016.

\bibitem{peng_iccv17}
X.~Peng, X.~Yu, K.~Sohn, D.~Metaxas, and M.~Chandraker.
\newblock Reconstruction for feature disentanglement in pose-invariant face
  recognition.
\newblock In {\em ICCV}, 2017.

\bibitem{dehaze_2016_eccv}
W.~Ren, S.~Liu, H.~Zhang, J.~Pan, X.~Cao, and M.-H. Yang.
\newblock Single image dehazing via multi-scale convolutional neural networks.
\newblock In {\em ECCV}, pages 154--169. Springer, 2016.

\bibitem{derain_cvpr17_video}
W.~Ren, J.~Tian, Z.~Han, A.~Chan, and Y.~Tang.
\newblock Video desnowing and deraining based on matrix decomposition.
\newblock In {\em Proceedings of the IEEE Conference on Computer Vision and
  Pattern Recognition}, pages 4210--4219, 2017.

\bibitem{unet}
O.~Ronneberger, P.~Fischer, and T.~Brox.
\newblock U-net: Convolutional networks for biomedical image segmentation.
\newblock In {\em International Conference on Medical Image Computing and
  Computer-Assisted Intervention}, pages 234--241. Springer, 2015.

\bibitem{derain_video_ijcv}
V.~Santhaseelan and V.~K. Asari.
\newblock Utilizing local phase information to remove rain from video.
\newblock {\em International Journal of Computer Vision}, 112(1):71--89, 2015.

\bibitem{vgg}
K.~Simonyan and A.~Zisserman.
\newblock Very deep convolutional networks for large-scale image recognition.
\newblock {\em arXiv preprint arXiv:1409.1556}, 2014.

\bibitem{crowd_counting_Vishwanath}
V.~A. Sindagi and V.~M. Patel.
\newblock Generating high-quality crowd density maps using contextual pyramid
  cnns.
\newblock In {\em ICCV}, 2017.

\bibitem{ssim}
Z.~Wang, A.~C. Bovik, H.~R. Sheikh, and E.~P. Simoncelli.
\newblock Image quality assessment: from error visibility to structural
  similarity.
\newblock {\em IEEE TIP}, 13(4):600--612, 2004.

\bibitem{derain_iccv17_video}
W.~Wei, L.~Yi, Q.~Xie, Q.~Zhao, D.~Meng, and Z.~Xu.
\newblock Should we encode rain streaks in video as deterministic or
  stochastic?
\newblock In {\em Proceedings of the IEEE Conference on Computer Vision and
  Pattern Recognition}, pages 2516--2525, 2017.

\bibitem{deep_supervision}
S.~Xie and Z.~Tu.
\newblock Holistically-nested edge detection.
\newblock In {\em Proceedings of the IEEE international conference on computer
  vision}, pages 1395--1403, 2015.

\bibitem{tao_stackgan_cvpr2018}
T.~Xu, P.~Zhang, Q.~Huang, H.~Zhang, Z.~Gan, X.~Huang, and X.~He.
\newblock Attngan: Fine-grained text to image generation with attentional
  generative adversarial networks.
\newblock In {\em CVPR}, 2018.

\bibitem{jia_differ}
J.~Xue, H.~Zhang, K.~Dana, and K.~Nishino.
\newblock Differential angular imaging for material recognition.
\newblock In {\em CVPR}, 2017.

\bibitem{derain_cvpr2017_multi}
W.~Yang, R.~T. Tan, J.~Feng, J.~Liu, Z.~Guo, and S.~Yan.
\newblock Deep joint rain detection and removal from a single image.
\newblock In {\em Proceedings of the IEEE Conference on Computer Vision and
  Pattern Recognition}, pages 1357--1366, 2017.

\bibitem{style_hang}
H.~Zhang and K.~Dana.
\newblock Multi-style generative network for real-time transfer.
\newblock {\em arXiv preprint arXiv:1703.06953}, 2017.

\bibitem{derain_csc_17}
H.~Zhang and V.~M. Patel.
\newblock Convolutional sparse and low-rank coding-based rain streak removal.
\newblock In {\em Applications of Computer Vision (WACV), 2017 IEEE Winter
  Conference on}, pages 1259--1267. IEEE, 2017.

\bibitem{derain_2017_zhang}
H.~Zhang, V.~Sindagi, and V.~M. Patel.
\newblock Image de-raining using a conditional generative adversarial network.
\newblock {\em arXiv preprint arXiv:1701.05957}, 2017.

\bibitem{dehaze_2017_joint}
H.~Zhang, V.~Sindagi, and V.~M. Patel.
\newblock Joint transmission map estimation and dehazing using deep networks.
\newblock {\em arXiv preprint arXiv:1708.00581}, 2017.

\bibitem{zizhao_mdnet}
Z.~Zhang, Y.~Xie, F.~Xing, M.~McGough, and L.~Yang.
\newblock Mdnet: A semantically and visually interpretable medical image
  diagnosis network.
\newblock In {\em CVPR}, 2017.

\bibitem{psp_net}
H.~Zhao, J.~Shi, X.~Qi, X.~Wang, and J.~Jia.
\newblock Pyramid scene parsing network.
\newblock In {\em Proceedings of the IEEE International Conference on Computer
  Vision}, pages 1--8, 2017.

\bibitem{feature_vis_bolei}
B.~Zhou, A.~Khosla, A.~Lapedriza, A.~Oliva, and A.~Torralba.
\newblock Learning deep features for discriminative localization.
\newblock In {\em Proceedings of the IEEE Conference on Computer Vision and
  Pattern Recognition}, pages 2921--2929, 2016.

\bibitem{derain_iccv17}
L.~Zhu, C.-W. Fu, D.~Lischinski, and P.-A. Heng.
\newblock Joint bi-layer optimization for single-image rain streak removal.
\newblock In {\em Proceedings of the IEEE international conference on computer
  vision}, pages 2526--2534, 2017.

\bibitem{yizhu_hiddentwo}
Y.~Zhu, Z.~Lan, S.~Newsam, and A.~G. Hauptmann.
\newblock Hidden two-stream convolutional networks for action recognition.
\newblock {\em arXiv preprint arXiv:1704.00389}, 2017.

\end{thebibliography}
}

\end{document}